\documentclass[sigconf]{acmart}
\usepackage{booktabs}
\usepackage{array}
\usepackage{multirow}
\usepackage{colortbl}
\usepackage{xcolor}
\usepackage{amssymb}
\usepackage{makecell}  % 添加这个包以支持\makecell
\usepackage{caption}   % 添加这个包以更好地控制表格标题
\usepackage{placeins}  % 提供 \FloatBarrier 命令，防止浮动体越界
\usepackage{cuted}

\usepackage{longtable} % 支持跨页长表格
\usepackage{multicol}  % 支持在单栏页面中局部双栏排版

\usepackage{seqsplit}
\makeatletter
\let\texttt@orig\texttt
\renewcommand{\texttt}[1]{\texttt@orig{\seqsplit{#1}}}
\newcommand{\codeinline}[1]{\mbox{\texttt@orig{#1}}}
\makeatother
\usepackage{pifont}    % 添加 pifont 包以使用 \ding 系列符号
\usepackage{tabularx}  % 添加 tabularx 包，使用 X 列类型（解决 tabular* 下 \rowcolor 不铺满的问题）
\usepackage{tikz}      % 用于绘制半透明进度条和自定义图标
\usepackage{pgf}       % tikz 依赖
\usetikzlibrary{calc}  % tikz 计算库
\usepackage{graphicx}  % 用于插入模型图标
\graphicspath{{./}{./figures/}{./figures/icons/}}  % 让 \includegraphics 能在不同工作目录下都找到图标

\hypersetup{hidelinks}

\definecolor{hl}{RGB}{253, 232, 215}
\definecolor{hlrow}{RGB}{226, 232, 240}      % 底部 Ours 行的浅蓝灰高亮带
\definecolor{markgreen}{RGB}{46, 160, 67}    % 绿色 ✓
\definecolor{markred}{RGB}{210, 50, 50}      % 红色 ✗
\definecolor{markorange}{RGB}{235, 140, 30}  % 橙色 ◇

\definecolor{barbg}{RGB}{243, 239, 250}
\definecolor{barfg}{RGB}{171, 148, 210}

\newcommand{\micon}[1]{\raisebox{-0.20\height}{\includegraphics[height=2.5ex]{#1}}}
\newcommand{\iconOpenAI}{\micon{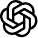}}
\newcommand{\iconClaude}{\micon{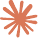}}
\newcommand{\iconGemini}{\micon{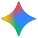}}
\newcommand{\iconDeepSeek}{\micon{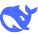}}
\newcommand{\iconKimi}{\micon{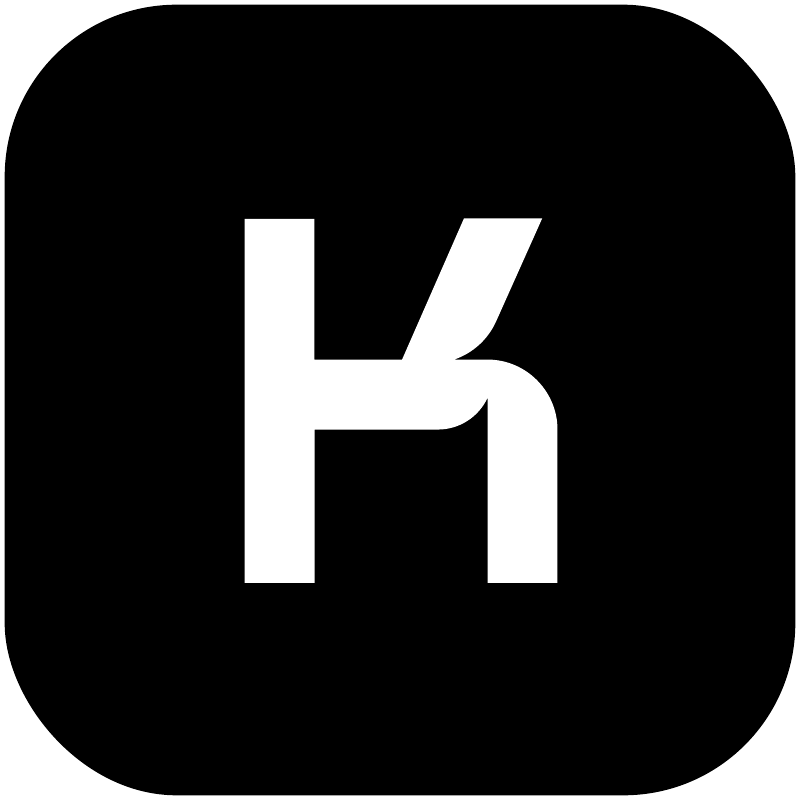}}
\newcommand{\iconMinimax}{\micon{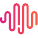}}
\newcommand{\iconHunyuan}{\micon{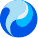}}
\newcommand{\iconGLM}{\micon{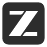}}

\newcommand{\scorebar}[1]{%
  \begin{tikzpicture}[baseline=(t.base), line width=0pt]
    \fill[barbg, rounded corners=1.0pt] (0,0.04) rectangle (1.0,0.37);
    \fill[barfg, rounded corners=1.0pt, fill opacity=0.70]
          (0,0.04) rectangle ({#1*0.010},0.37);
    \node[anchor=base, inner sep=0.2pt] (t) at (0.5, 0.150) {#1};
  \end{tikzpicture}%
}

\definecolor{barfgbest}{RGB}{124, 96, 184}
\newcommand{\scorebarbest}[1]{%
  \begin{tikzpicture}[baseline=(t.base), line width=0pt]
    \fill[barbg, rounded corners=1.0pt] (0,0.04) rectangle (1.0,0.37);
    \fill[barfgbest, rounded corners=1.0pt, fill opacity=0.90]
          (0,0.04) rectangle ({#1*0.010},0.37);
    \node[anchor=base, font=\bfseries, inner sep=0.2pt] (t) at (0.5, 0.150) {#1};
  \end{tikzpicture}%
}

\newcommand{\cm}{{\color{markgreen}\ding{51}}}
\newcommand{\xx}{{\color{markred}\ding{55}}}

\AtBeginDocument{%
  }

\acmConference[KDD '27]{Proceedings of the 33rd ACM SIGKDD Conference on
  Knowledge Discovery and Data Mining}{August 1--5, 2027}{San Jose, CA, USA}
\setcopyright{none}
\renewcommand\footnotetextcopyrightpermission[1]{}

\begin{document}

\title{DataClawEval: A  Benchmark for Data Engineering Agents in Real Industrial Harness}

\author{Debin Meng}
\authornote{Both authors contributed equally to this research.}
\email{debinmeng@tencent.com}
\affiliation{%
  \institution{Tencent}
  \city{Shenzhen}
  \country{China}}
\affiliation{%
  \institution{Xidian University}
  \city{Xi'an}
  \country{China}}

\author{Jiaming Yang}
\authornotemark[1]
\email{besmingyang@tencent.com}
\affiliation{%
  \institution{Tencent}
  \city{Beijing}
  \country{China}}
\affiliation{%
  \institution{Sun Yat-Sun University}
  \city{Guangzhou}
  \country{China}}

\author{Zefang Zong}
\authornote{Corresponding author.}
\email{willzong@tencent.com}
\affiliation{%
  \institution{Tencent}
  \city{Beijing}
  \country{China}}

\author{Tengyue Xu}
\email{leooxu@tencent.com}
\affiliation{%
  \institution{Tencent}
  \city{Beijing}
  \country{China}}
\affiliation{%
  \institution{South China Normal University}
  \city{Guangzhou}
  \country{China}}

\author{Haining Xie}
\email{hainingxie@tencent.com}
\affiliation{%
  \institution{Tencent}
  \city{Beijing}
  \country{China}}

\author{Yang Li}
\email{thomasyngli@tencent.com}
\affiliation{%
  \institution{Tencent}
  \city{Beijing}
  \country{China}}

\author{Peng Chen}
\email{pengchen@tencent.com}
\affiliation{%
  \institution{Tencent}
  \city{Shenzhen}
  \country{China}}

\renewcommand{\shortauthors}{Meng and Yang et al.}

\begin{abstract}
Large language models (LLMs) and LLM-based agents are increasingly being deployed to automate complex workflows, promising to revolutionize data management and processing. However, existing benchmarks predominantly focus on simplified Text-to-SQL translation or data analysis, leaving the critical and complex domain of end-to-end data engineering largely unexplored. To bridge this gap, we introduce DataClawEval, the first comprehensive benchmark designed specifically to evaluate the end-to-end task completion capabilities of autonomous agents in real-world data engineering scenarios. Built upon production-grade code authored by professional enterprise data engineers, it comprises 100 rigorous, end-to-end tasks spanning five execution engines: PySpark, MySQL, HiveSQL, PrestoSQL/Trino, and FlinkSQL. Rather than non-deterministic LLM-as-a-judge scoring, each task is executed within a case-specific, isolated sandbox and graded by deterministic, rule-based scripts. Evaluating 16 frontier agents exposes critical limitations: The strongest model attains only 74.9 overall, and no single model dominates, as each excels on a different engine, revealing strict domain specialization rather than omnipotent proficiency. Thus, autonomous data engineering remains a formidable, unresolved challenge. We release our dataset, containerized environments, and deterministic evaluation scripts at \url{https://github.com/Dicemy/DataClawEval/tree/master}.
\end{abstract}

\begin{CCSXML}
<ccs2012>
   <concept>
       <concept_id>10010147.10010178.10010219.10010221</concept_id>
       <concept_desc>Computing methodologies~Intelligent agents</concept_desc>
       <concept_significance>500</concept_significance>
       </concept>
   <concept>
       <concept_id>10002951.10002952.10003190</concept_id>
       <concept_desc>Information systems~Database management system engines</concept_desc>
       <concept_significance>300</concept_significance>
       </concept>
   <concept>
       <concept_id>10010147.10010178.10010179</concept_id>
       <concept_desc>Computing methodologies~Natural language processing</concept_desc>
       <concept_significance>300</concept_significance>
       </concept>
 </ccs2012>
\end{CCSXML}

\ccsdesc[500]{Computing methodologies~Intelligent agents}
\ccsdesc[300]{Information systems~Database management system engines}
\ccsdesc[300]{Computing methodologies~Natural language processing}

\keywords{Data Engineering, Autonomous Agent, Enterprise Data Systems}

\maketitle
\section{Introduction}

Recent advances in large language models (LLMs) have rapidly transformed them from conversational assistants into autonomous agents capable of solving complex tasks through iterative reasoning, tool invocation, code execution, and interaction with external environments. General-purpose agent systems such as Tencent CodeBuddy~\cite{tencent2025codebuddy}, Claude Code~\cite{anthropic2025claudecode}, OpenAI Codex~\cite{chen2021evaluating}, and OpenClaw~\cite{steinberger2025openclaw} have demonstrated impressive performance in realistic software engineering workflows, while benchmarks such as WildClawBench~\cite{ding2026wildclawbench} and ClawEval~\cite{ye2026claweval} have significantly accelerated research on autonomous agents in general task execution. As these capabilities continue to mature, autonomous agents are naturally extending beyond software engineering to enterprise data engineering, giving rise to data engineering agents that assist developers in building ETL pipelines, implementing batch and streaming data processing jobs, debugging execution failures, and validating production workflows.

\begin{table*}[t]
\centering
\small
\renewcommand{\arraystretch}{1.3}
% 使用 tabularx 让 8 列表格在双栏页面里撑满整页 \textwidth，
% 避免列宽被自动压到单栏宽度导致表头换行拥挤。
% 注意：不能用 tabular* + \extracolsep{\fill} 拉伸——\fill 弹性胶在列之间，
% 不属于任何单元格，\rowcolor 涂不到，高亮带会在第 1、2 列之间断开。
% 改用 tabularx 的 X 列：多余宽度在单元格内部吸收，行间无未上色胶缝，
% 配合首尾 @{}，\rowcolor 可从最左连续铺到最右。
\begin{tabularx}{\textwidth}{@{}Xccccccc@{}}
\toprule
\textbf{Benchmark}
  & \textbf{\makecell{No Schema\\Prior}}
  & \textbf{\makecell{Cross\\Domain}}
  & \textbf{\makecell{Process\\Evaluation}}
  & \textbf{\makecell{Real-World\\Scenario}}
  & \textbf{\makecell{Data\\Engineering}}
  & \textbf{\makecell{Multi SQL\\Engine}}
  & \textbf{\makecell{Fully Rule-\\based Eval}} \\
\midrule
BIRD \cite{li2023bird}                           & \xx & \cm & \xx & \cm & \xx & \xx & \cm \\
Spider 2.0 \cite{lei2025spider2}                 & \xx & \cm & \xx & \cm & \xx & \cm & \cm \\
InfiAgent-DABench \cite{hu2024infiagent}         & \xx & \cm & \xx & \xx & \xx & \xx & \cm \\
DA-Code \cite{huang2024dacode}                   & \xx & \cm & \xx & \cm & \cm & \xx & \cm \\
DABstep \cite{egg2025dabstep}                    & \cm & \cm & \xx & \cm & \xx & \xx & \cm \\
FDABench \cite{wang2026fdabench}                 & \cm & \cm & \xx & \xx & \xx & \xx & \xx \\
FinanceBench \cite{islam2023financebench}        & \cm & \xx & \xx & \cm & \xx & \xx & \xx \\
DataClawBench \cite{zhang2025dataclawbench}      & \cm & \cm & \cm & \cm & \xx & \xx & \xx \\
\rowcolor{hlrow}
\textbf{Ours (DataClawEval)}                     & \cm & \cm & \cm & \cm & \cm & \cm & \cm \\
\bottomrule
\end{tabularx}
\caption{
Comparison of DataClawEval against eight representative data-related agent benchmarks across seven key capability dimensions.
\cm\ indicates the benchmark supports the property; \xx\ indicates it does not.
}
\label{tab:benchmark-comparison}
\end{table*}

Enterprise data engineering, however, differs fundamentally from conventional coding or Text-to-SQL tasks. Given a natural-language requirement, a data engineer must inspect database schemas, understand business semantics, identify upstream data sources, reason about transformation logic, implement executable programs using heterogeneous data-processing technologies, iteratively debug runtime failures, and finally validate generated outputs. Such workflows span multiple execution engines, including PySpark~\cite{zaharia2012resilient}, HiveSQL~\cite{thusoo2010hive}, MySQL~\cite{oracle_mysql}, PrestoSQL/Trino~\cite{sethi2019presto}, and FlinkSQL~\cite{carbone2015flink}, and require agents to jointly perform reasoning, programming, execution, debugging, and environment interaction. Consequently, evaluating data engineering agents requires substantially more than measuring code generation quality or SQL correctness.

Despite the rapid emergence of data-oriented agent benchmarks, none adequately captures the characteristics of real-world enterprise data engineering; existing efforts fall into three families that each isolate a single capability. \textbf{(i) Text-to-SQL.} BIRD~\cite{li2023bird} and Spider 2.0~\cite{lei2025spider2} evaluate the synthesis of a SQL answer from a natural-language question over a fixed schema, scored by execution accuracy: BIRD stresses large, dirty databases (12{,}751 question--SQL pairs over 95 real databases), while Spider 2.0 raises the difficulty to enterprise cloud warehouses with long schemas and multiple dialects. Even so, the objective remains predominantly single-step query generation rather than building and running an end-to-end pipeline. \textbf{(ii) Code-execution data analysis.} InfiAgent-DABench~\cite{hu2024infiagent}, DA-Code~\cite{huang2024dacode}, and DABstep~\cite{egg2025dabstep} require an agent to write and iteratively execute code (Python/SQL/Bash) in a sandbox to reach an analytical answer---closed-form statistics over CSV files, data-wrangling/ML tasks, or multi-step factoid reasoning over financial records---so success is measured by whether a final answer matches, not whether a deployable data-processing job is correctly engineered. \textbf{(iii) Heterogeneous and exploratory analysis.} FDABench~\cite{wang2026fdabench} shifts the emphasis to retrieving and reasoning across structured databases, unstructured documents, web, and multimodal content, whereas DataClawBench~\cite{zhang2025dataclawbench} targets exploratory analysis under limited schema priors over noisy, cross-domain financial records. The lone step toward engineering, ELT-Bench~\cite{jin2025eltbench}, builds ELT pipelines but is confined to a single dbt/Airbyte-on-warehouse stack. Across all three families, tasks are designed for analysis, reasoning, or query/answer generation rather than enterprise data engineering: they generally lack realistic ETL engineering tasks, heterogeneous batch and streaming execution engines, reproducible runtime environments, and deterministic protocols that verify the complete engineering process rather than only a final answer. As a result, the research community still lacks a benchmark for systematically evaluating autonomous agents on realistic enterprise data engineering tasks (Table~\ref{tab:benchmark-comparison}).

To bridge this gap, we present \textbf{DataClawEval}, the first executable benchmark specifically designed for autonomous data engineering agents. \textbf{First}, we curate a production-grounded benchmark comprising 100 end-to-end data development tasks across PySpark, HiveSQL, MySQL, PrestoSQL/Trino, and FlinkSQL, covering both batch and streaming workloads in five representative business domains.
\textbf{Second}, we present a human-in-the-loop methodology for constructing answer-identifiable benchmark tasks from production code. Our pipeline cleans and filters source artifacts, reconstructs task intents and inputs, and validates task discriminability and grader reliability through execution checks, expert perturbations, score-based verification, and final expert audit. It further enables deterministic evaluation of both final artifacts and development processes using case-specific grading programs in isolated sandbox environments.
\textbf{Third}, we provide the first comprehensive evaluation of 16 frontier LLM-based data engineering agents on DataClawEval. Our analysis reveals substantial headroom, highly uneven engine difficulty, no universally dominant model, and no positive relationship between resource consumption and solution quality.

Our contributions are summarized as follows:

\begin{itemize}
    \item We introduce \textbf{DataClawEval}, the first executable benchmark grounded in real-world commercial data sources and workflows, enabling faithful evaluation of autonomous agents in realistic enterprise data engineering scenarios beyond existing coding, SQL, and data-analysis benchmarks.

    \item We develop a human-in-the-loop data construction pipeline for building answer-identifiable tasks and an end-to-end evaluation framework, featuring reproducible isolated sandboxes and case-specific deterministic grading scripts for trustworthy, fine-grained, and reproducible assessment.

    \item We conduct a controlled evaluation of 16 frontier LLMs under a unified agent harness across 100 tasks and five data engines, showing that engine difficulty is highly uneven, no single model dominates across all engines, and token cost does not track solution quality.
\end{itemize}

\section{Preliminaries} \label{sec:preliminary}

We first describe the data sources and task background of DataClawEval, covering the scope of the enterprise engineering workflow that each task must reproduce.

\subsection{Task Scope} \label{sec:task-scope}

Each task in DataClawEval corresponds to a realistic data engineering request derived from production-grade code authored by professional enterprise data engineers. The benchmark includes both offline computation and online computation workloads. Offline tasks require the agent to build batch transformations using PySpark, HiveSQL, MySQL, or PrestoSQL/Trino. Online tasks focus on streaming-oriented FlinkSQL development, where the agent must define source tables, sink tables, event-time attributes, watermarks, windows, and streaming aggregations. The current task suite is organized as shown in Figure~\ref{fig:task-dist} (see Appendix~\ref{sec:task-listing} for the full per-task listing). To reflect realistic industrial workloads, tasks are drawn from five business domains, so that the benchmark measures not only syntactic correctness but also domain-grounded analytical reasoning:

\begin{itemize}
    \item \textbf{Ops \& Resource Governance} (30 tasks): GPU/notebook instance
    governance, message-queue and quota management, resource-cost aggregation,
    and inference-service capacity planning.
    \item \textbf{Data Analytics \& User Growth} (30 tasks): user activity and
    retention, PV/UV statistics, real-time windowed metrics, and funnel/TopN
    analysis over streams and batch tables.
    \item \textbf{Security \& Risk Control} (16 tasks): security scanning,
    threat-intelligence correlation, anomaly detection, and risk scoring.
    \item \textbf{Content, Community \& Dev-Efficiency} (14 tasks): content
    quality scoring, community wide-table modeling, and code review analytics, and engineering efficiency analytics.
    \item \textbf{Advertising \& Marketing} (10 tasks): ad attribution,
    campaign tagging, and marketing-effect aggregation.
\end{itemize}

\begin{figure}[t]
\centering
\includegraphics[width=\linewidth]{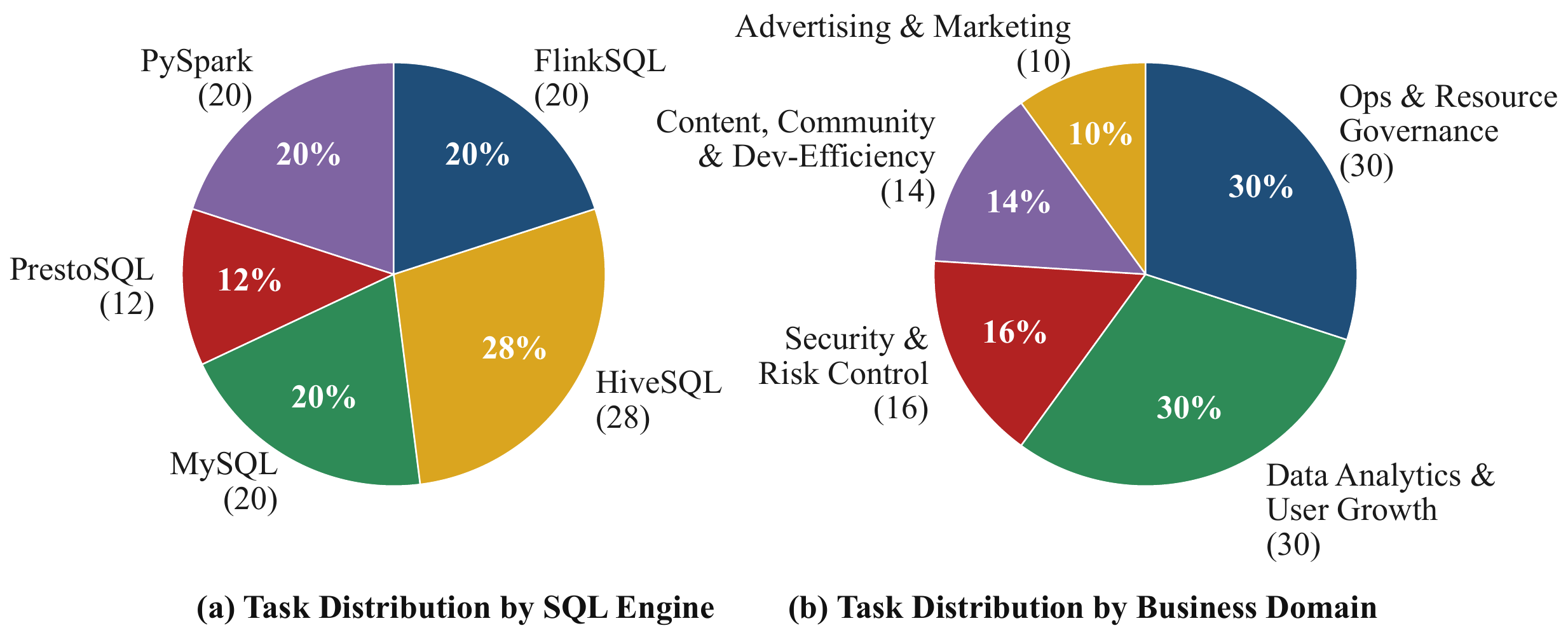}
\caption{Composition of the 100-task DataClawEval suite:
\textbf{(a)} distribution by execution engine and \textbf{(b)} distribution by business domain.}
\label{fig:task-dist}
\end{figure}

\subsection{Realistic Data Engineering Workflow} \label{sec:workflow}

The tasks are designed to reproduce the workflow of a human data engineer. A typical task requires the agent to:

\begin{enumerate} \item read a natural language requirement; \item identify input and output tables from the prompt; \item connect to the corresponding execution engine; \item inspect table schemas and sample rows; \item infer transformation rules, join keys, partition fields, timestamp semantics, and boundary conditions; \item write executable SQL or Python code; \item run the code in the sandbox; \item inspect runtime errors and revise the implementation; \item validate the produced output table or file; \item finish with a materialized result artifact. \end{enumerate}

This workflow reflects real-world data engineering practice, where engineers often work without a complete formal schema or reference query and must inspect data, explore the execution environment, and iteratively refine implementations based on execution feedback. DataClawEval reproduces this process to assess agents' practical, end-to-end data engineering capabilities. Concretely, a single task often chains together schema inference, multi-table joins, partition and timestamp handling, boundary-condition reasoning, and iterative debugging, exposing the compounding difficulty that agents must overcome to reach a correct materialized result. We illustrate this with five representative case studies, one per business domain and collectively spanning all five execution engines, ranging from replica-count forecasting for autoscaling inference services to interval joins with windowed aggregation over order-payment streams (see Appendix~\ref{sec:case-studies}).

\section{DataClawEval} \label{sec:benchmark}

DataClawEval is designed to measure whether an agent can autonomously complete realistic data engineering tasks in a live execution environment. Unlike static benchmarks, where the model only produces an answer, DataClawEval requires the agent to interact with a running sandbox. The sandbox contains task-specific data, database services, execution engines, and command-line tools. The agent must use these resources to produce an executable solution and materialize the required result. The remainder of this section details the data construction pipeline (Section~\ref{sec:data-curation}) and the evaluation framework (Section~\ref{sec:evaluation-framework}).

\begin{figure*}[!t] % 使用 figure* 使得图片横跨双栏，[!t] 通常能让它更好地排版在页面顶部
  \centering
  % 将宽度从 0.5\linewidth 改为 1.0\textwidth（或 0.95\textwidth 留出微调边距）
  \includegraphics[width=1.0\textwidth]{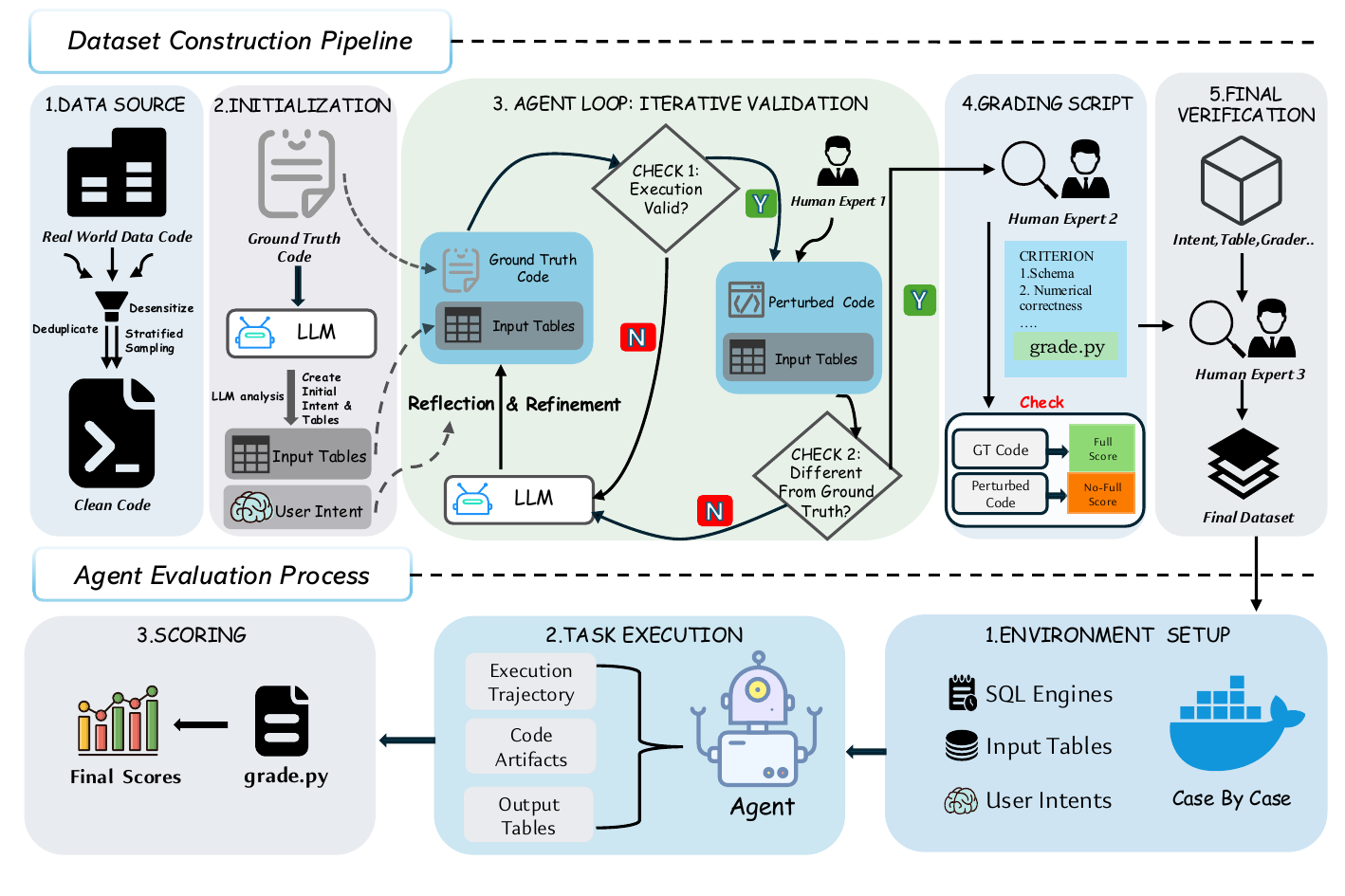}
  \caption{Overview of DataClawEval's two coupled pipelines.
  \textbf{Top (Dataset Construction):} real data engineering code is cleaned and sampled to create benchmark tasks. An LLM synthesizes user intents and input tables, which are iteratively validated through agent-in-the-loop execution and expert perturbations. Case-specific graders and domain experts then verify task correctness, discriminability, and quality.
  \textbf{Bottom (Agent Evaluation):} For each case, an agent operates in an isolated Docker environment to generate executable code and output tables. A case-specific grader evaluates the resulting artifacts and assigns the final score.}
  \label{fig:overview}
\end{figure*}

\subsection{Dataset Construction Pipeline}
\label{sec:data-curation}

Constructing realistic benchmark tasks from production data engineering pipelines involves more than collecting authentic code artifacts. A high-quality benchmark requires every task to establish a consistent correspondence between the user intent, synthesized input tables, ground-truth implementation, and evaluation protocol, so that correct solutions are executable, incorrect solutions are distinguishable, and agent performance can be assessed objectively. To achieve this goal, we design a data construction pipeline that combines LLM-assisted task reconstruction with iterative human-in-the-loop validation. As illustrated in Figure~\ref{fig:overview}, the pipeline progressively reconstructs missing task specifications, validates the discriminative power of synthesized inputs, and verifies task-specific grading scripts, producing realistic, answer-identifiable, and reproducible benchmark instances.

\noindent\textbf{Stage 1: Curating high-quality production code.}
We begin by collecting real production data engineering code, removing sensitive information and low-quality duplicates, and applying stratified sampling to preserve diversity across programming paradigms, SQL dialects, and business scenarios. The retained implementations serve as the ground-truth code, ensuring that every benchmark task originates from authentic engineering practice rather than synthetic generation.

\noindent\textbf{Stage 2: Reconstructing executable task specifications.}
Production repositories typically contain only implementation code, while the original data engineering requests and testing data are unavailable. Starting from the ground-truth implementation, we employ an LLM to infer the underlying user intent using the code together with the associated table schemas. The LLM then synthesizes representative input tables that satisfy the inferred intent and execution requirements, producing an initial task specification consisting of the natural-language request, input tables, and ground-truth implementation. Since this initialization is generated automatically, it may still omit important semantic constraints or fail to distinguish the correct implementation from plausible alternatives, motivating the subsequent validation stage.

\noindent\textbf{Stage 3: Iterative validation for answer-identifiable task construction.}
The core of our pipeline is an iterative validation loop whose objective is to ensure that the synthesized input tables uniquely characterize the ground-truth implementation rather than merely supporting successful execution. First, the generated input tables are executed together with the ground-truth code to verify that the reconstructed task is executable and produces valid outputs. However, successful execution alone is insufficient because many incomplete or ambiguous inputs may still allow incorrect implementations to generate identical results. We therefore introduce a second validation stage inspired by \textbf{differential testing}~\cite{mckeeman1998differential}. Human experts intentionally perturb the ground-truth implementation while preserving its executability, and the synthesized input tables are required to produce outputs that differ from those of the original implementation. If either execution fails or the perturbed implementation remains indistinguishable from the ground truth, the task specification is considered insufficiently discriminative. The resulting feedback is returned to the LLM, which revises the reconstructed intent, adjusts table schemas, and modifies the synthesized records until the input tables satisfy two complementary requirements: they enable the ground-truth implementation to produce valid outputs, while reliably distinguishing it from semantically incorrect implementations.

\noindent\textbf{Stage 4: Developing and validating task-specific grading scripts.}
After the task specification has been validated, human experts develop a dedicated grading script for each benchmark instance. Rather than relying on generic output matching, every grading program is customized according to the task semantics, covering aspects such as schema correctness, numerical correctness, business logic, and other task-specific constraints. The grading script is subsequently validated against both the ground-truth implementation and the perturbed implementation. Specifically, the ground-truth code must consistently receive a full score, whereas the perturbed implementation must fail to obtain full marks. This second layer of verification complements the execution-based validation by ensuring that the evaluation protocol itself faithfully distinguishes correct solutions from semantically incorrect ones.

\noindent\textbf{Stage 5: Final expert verification.}
Finally, every benchmark instance undergoes a comprehensive manual audit. Domain experts jointly inspect the reconstructed user intent, synthesized input tables, ground-truth implementation, grading script, and expected outputs to ensure their overall consistency. Together, the execution-based validation and grading verification provide two complementary safeguards: the synthesized inputs must distinguish the ground-truth implementation from incorrect alternatives, and the grading script must assign full credit to the ground-truth implementation and any semantically equivalent solution. Consequently, every benchmark instance establishes a reliable correspondence between the user intent, input tables, ground-truth implementation, and evaluation protocol, making it suitable as an objective, discriminative, and reproducible benchmark task.

\subsection{Evaluation Framework}
\label{sec:evaluation-framework}

Evaluating autonomous data engineering agents presents two fundamental challenges. First, the execution environment must faithfully reproduce realistic engineering settings while preventing information leakage between benchmark cases. Second, the evaluation should measure not only the correctness of the final artifacts but also the quality of the engineering process, since practical data engineering heavily relies on iterative exploration, schema inspection, and result validation. To address these challenges, DataClawEval adopts an isolated, case-by-case evaluation framework that jointly evaluates execution outcomes and engineering behaviors.

\noindent\textbf{Isolated task execution.}
Each benchmark instance is executed inside a fresh Docker container with all required services (data warehouses, query engines, metadata services, and execution gateways) pre-initialized. The workspace is sanitized to contain only the task specification and engineering utilities, forcing the agent to independently explore table schemas, understand data distributions, and construct executable programs. After execution, a task-specific evaluator performs automatic assessment inside the same container, and all artifacts are collected for aggregation. This containerized design guarantees reproducibility and eliminates inconsistencies from heterogeneous runtime environments.

\noindent\textbf{Artifact-oriented evaluation.}
Unlike conventional code generation benchmarks that compare source code directly, DataClawEval evaluates the correctness of the produced data artifacts. Since multiple implementations may correctly solve the same business requirement, comparing generated code against a reference implementation would unnecessarily penalize valid alternative solutions.

Instead, each benchmark case defines its own evaluation script that validates the generated outputs against business-specific correctness criteria. Although the exact metrics vary across tasks, the artifact score generally considers five complementary aspects:
\begin{itemize}
    \item \textbf{Executability}, ensuring that the submitted program successfully produces the expected outputs;
    \item \textbf{Schema correctness}, including table names, column definitions, data types, and storage formats;
    \item \textbf{Row-level alignment}, measuring the completeness and correctness of generated records;
    \item \textbf{Numerical accuracy}, which compares key quantitative fields specified by each task;
    \item \textbf{Categorical correctness}, which validates business labels and other discrete attributes.
\end{itemize}

\noindent\textbf{Process-oriented evaluation.}
Correct outputs alone provide an incomplete picture of an agent's capability. In realistic data engineering workflows, experienced engineers routinely inspect schemas, sample source tables, verify intermediate results, and iteratively refine their implementations. These behaviors are essential for solving complex analytical tasks but remain invisible under output-only evaluation.

Therefore, DataClawEval additionally evaluates the execution trajectory of each agent. The process score measures three aspects of engineering quality: exploration adequacy, execution efficiency, and self-verification. Exploration adequacy assesses whether the agent sufficiently investigates the available data before implementation. Execution efficiency evaluates whether the solution is obtained with a reasonable number of execution attempts while avoiding unnecessary retries and redundant operations. Self-verification measures whether the agent actively validates the generated outputs through post-execution inspection, such as checking row counts, aggregation results, or critical numerical statistics.

\noindent\textbf{Overall score.}
The final score is computed as a weighted combination of the artifact score and the process score,

\begin{equation}
\mathrm{Score}
=
\alpha S_{\mathrm{artifact}}
+
(1-\alpha)S_{\mathrm{process}},
\end{equation}

where $\alpha$ controls the relative importance of final correctness and engineering quality. In most benchmark cases, we set $\alpha=0.7$, placing greater emphasis on the correctness of the produced artifacts while still rewarding efficient and reliable engineering behaviors. Task designers may adjust $\alpha$ for scenarios in which execution quality plays a more critical role.

\section{Experiments} \label{sec:experiments}
\begin{table*}[t]
\centering
\small
\caption{Overall and engine-specific performance of 16 LLM-based agents on DataClawEval. Scores are reported on a 0--100 scale, with bars indicating relative performance; token usage is averaged per task and reported in thousands (k). Higher scores and lower token usage are better, and the best result in each column is shown in \textbf{bold}.}
\label{tab:main-results}
\renewcommand{\arraystretch}{1.45}
\setlength{\tabcolsep}{2.4pt}
\resizebox{\textwidth}{!}{%
\begin{tabular}{l ccccccccccccc}
\toprule
 & \multicolumn{2}{c}{Overall} & \multicolumn{2}{c}{PySpark} & \multicolumn{2}{c}{HiveSQL} & \multicolumn{2}{c}{MySQL} & \multicolumn{2}{c}{PrestoSQL/Trino} & \multicolumn{2}{c}{FlinkSQL} \\
\cmidrule(lr){2-3}\cmidrule(lr){4-5}\cmidrule(lr){6-7}\cmidrule(lr){8-9}\cmidrule(lr){10-11}\cmidrule(lr){12-13}
Model & Score & Tokens & Score & Tokens & Score & Tokens & Score & Tokens & Score & Tokens & Score & Tokens \\
\midrule
\iconOpenAI\ GPT 5.5~\cite{openai2026gpt55}
  & \scorebarbest{74.9} & 299.8k & \scorebar{73.9} & 387.1k & \scorebarbest{69.8} & 294.0k & \scorebarbest{88.8} & 322.2k & \scorebar{75.3} & 296.9k & \scorebar{69.0} & 199.7k \\
\iconClaude\ Claude Opus 4.8~\cite{anthropic2026opus48}
  & \scorebar{74.3} & 318.3k & \scorebarbest{83.8} & 319.4k & \scorebar{65.0} & 299.0k & \scorebar{85.1} & 399.1k & \scorebar{74.7} & 320.4k & \scorebar{66.9} & 262.3k \\
\iconClaude\ Claude Sonnet 5~\cite{anthropic2026sonnet5}
  & \scorebar{73.8} & 457.9k & \scorebar{79.1} & 515.6k & \scorebar{58.6} & 344.2k & \scorebar{84.2} & 748.2k & \scorebar{66.3} & 299.3k & \scorebar{83.9} & 364.3k \\
\iconGemini\ Gemini 3.1 Pro~\cite{google2026gemini31pro}
  & \scorebar{73.7} & 292.4k & \scorebar{73.2} & 299.1k & \scorebar{63.6} & 313.9k & \scorebar{79.3} & \textbf{318.8k} & \scorebar{72.8} & 371.3k & \scorebar{83.2} & 181.9k \\
\iconGemini\ Gemini 3.5 Flash~\cite{google2026gemini35flash}
  & \scorebar{73.3} & 973.8k & \scorebar{75.3} & 1152.1k & \scorebar{62.9} & 758.1k & \scorebar{85.2} & 1040.4k & \scorebarbest{77.7} & 1529.0k & \scorebar{71.7} & 697.7k \\
\iconDeepSeek\ DeepSeek V4 Flash~\cite{deepseek2026v4}
  & \scorebar{73.0} & 419.6k & \scorebar{66.4} & 380.1k & \scorebar{64.1} & 366.2k & \scorebar{79.1} & 627.2k & \scorebar{74.4} & 405.3k & \scorebarbest{85.0} & 334.7k \\
\iconMinimax\ MiniMax M3~\cite{minimax2026m3}
  & \scorebar{71.8} & 714.1k & \scorebar{67.5} & 634.6k & \scorebar{62.0} & 567.1k & \scorebar{81.6} & 1326.0k & \scorebar{75.4} & 572.4k & \scorebar{78.1} & 472.7k \\
\iconGLM\ GLM 5.1~\cite{glm2026glm5}
  & \scorebar{71.6} & 355.4k & \scorebar{80.3} & 380.2k & \scorebar{59.5} & 231.4k & \scorebar{79.1} & 457.3k & \scorebar{75.9} & 459.2k & \scorebar{69.7} & 339.2k \\
\iconDeepSeek\ DeepSeek V4 Pro~\cite{deepseek2026v4}
  & \scorebar{70.6} & 359.3k & \scorebar{61.7} & 351.6k & \scorebar{60.4} & 374.8k & \scorebar{83.7} & 476.0k & \scorebarbest{77.7} & 402.8k & \scorebar{76.5} & 202.5k \\
\iconKimi\ Kimi K2.6~\cite{kimi2025k2}
  & \scorebar{69.0} & 428.8k & \scorebar{63.2} & 324.1k & \scorebar{61.5} & 282.1k & \scorebar{83.6} & 764.4k & \scorebar{76.2} & 503.8k & \scorebar{66.3} & 358.2k \\
\iconGLM\ GLM 5.2~\cite{glm2026glm5}
  & \scorebar{68.8} & 403.9k & \scorebar{75.1} & 446.5k & \scorebar{49.6} & \textbf{225.5k} & \scorebar{83.0} & 677.6k & \scorebar{65.4} & 363.1k & \scorebar{77.3} & 361.7k \\
\iconKimi\ Kimi K2.7~\cite{kimi2025k2}
  & \scorebar{68.1} & 407.5k & \scorebar{70.3} & 486.4k & \scorebar{59.3} & 261.0k & \scorebar{75.4} & 522.9k & \scorebar{76.1} & 457.2k & \scorebar{66.1} & 388.5k \\
\iconOpenAI\ GPT 5.3 Codex~\cite{openai2026gpt53codex}
  & \scorebar{66.4} & \textbf{271.4k} & \scorebar{54.1} & 281.2k & \scorebar{67.1} & 274.0k & \scorebar{81.6} & 339.2k & \scorebar{69.2} & 288.4k & \scorebar{60.8} & \textbf{180.0k} \\
\iconHunyuan\ Hy3~\cite{tencent2026hy3}
  & \scorebar{66.0} & 468.7k & \scorebar{74.9} & 612.0k & \scorebar{62.8} & 356.2k & \scorebar{81.9} & 495.7k & \scorebar{67.2} & 410.9k & \scorebar{44.9} & 491.5k \\
\iconMinimax\ MiniMax M2.7~\cite{minimax2026m2}
  & \scorebar{63.7} & 501.0k & \scorebar{59.2} & 658.7k & \scorebar{63.6} & 424.7k & \scorebar{76.8} & 687.9k & \scorebar{66.4} & 353.2k & \scorebar{53.6} & 352.2k \\
\iconGLM\ GLM 5V Turbo~\cite{glm2026glm5v}
  & \scorebar{60.3} & 287.8k & \scorebar{48.0} & \textbf{229.4k} & \scorebar{57.7} & 319.1k & \scorebar{74.4} & 442.0k & \scorebar{64.8} & \textbf{187.4k} & \scorebar{59.4} & 208.7k \\
\bottomrule
\end{tabular}%
}
\end{table*}
\subsection{Settings} \label{sec:settings}

For each combination of model and task, we conduct a single run in an isolated Docker container with deterministic initialization. During the run, the agent receives only the task description and access to the live execution environment. The reference implementation and grader for the task are withheld until the run has ended. Each run is subject to a predefined time limit. The benchmark contains 100 task descriptions, evenly divided between English and Chinese (50 each; see Appendix~\ref{sec:bilingual}).

To ensure a controlled comparison, we evaluate all models using the same Tencent CodeBuddy agent scaffold~\cite{tencent2025codebuddy}, which provides a uniform interface for tool use, execution, and environment interaction. We keep the scaffold fixed throughout the evaluation and vary only the underlying LLM, so that performance differences primarily reflect model capability rather than differences in agent implementation.

The 16 evaluated models span eight model families: GPT 5.5~\cite{openai2026gpt55} and GPT 5.3 Codex~\cite{openai2026gpt53codex}; Claude Opus 4.8~\cite{anthropic2026opus48} and Claude Sonnet 5~\cite{anthropic2026sonnet5}; Gemini 3.1 Pro~\cite{google2026gemini31pro} and Gemini 3.5 Flash~\cite{google2026gemini35flash}; DeepSeek V4 Flash and DeepSeek V4 Pro~\cite{deepseek2026v4}; GLM 5.1 and GLM 5.2~\cite{glm2026glm5}, and GLM 5V Turbo~\cite{glm2026glm5v}; Kimi K2.6 and Kimi K2.7~\cite{kimi2025k2}; MiniMax M3~\cite{minimax2026m3} and MiniMax M2.7~\cite{minimax2026m2}; and Hy3~\cite{tencent2026hy3}.

Our primary metric is the mean overall score across tasks. Table~\ref{tab:main-results} reports the mean score and token usage both across the full benchmark and separately for each execution engine, enabling comparison of aggregate performance, engine-specific capabilities, and computational cost.

\subsection{Main Results} \label{sec:main-results}

Table~\ref{tab:main-results} reports the main benchmark results for 16 agents on DataClawEval, broken down by engine. The results demonstrate that DataClawEval offers substantial headroom and remains far from saturated: even the strongest model, GPT 5.5, attains an average score of only $74.9$, leaving ample room to distinguish future, more capable agents. At the same time, the benchmark provides a rich and discriminative signal: aggregate scores span a wide $60.3$--$74.9$ range across models, and, as we show below, the per-engine breakdown reveals pronounced differences that a single leaderboard number would obscure. Together these properties indicate that DataClawEval is well calibrated for the current generation of data engineering agents: challenging enough to avoid ceiling effects, yet fine-grained enough to expose meaningful capability gaps.

\noindent\textbf{Engine difficulty is highly uneven.}
Performance varies far more across engines than across models. MySQL is by far the easiest engine; every agent scores above $74$, and the top model reaches $88.8$, likely because relational SQL is heavily represented in pretraining data. HiveSQL is consistently the hardest: no agent exceeds $69.8$ and several fall below $60$ (e.g., GLM 5.2 at $49.6$), reflecting the difficulty of Hive-specific dialect and semantics. PySpark and FlinkSQL show the largest spread across models ($48.0$--$83.8$ and $44.9$--$85.0$, respectively), suggesting that these engines sharply separate capable agents from weaker ones.

\noindent\textbf{No single model dominates all engines.}
While GPT 5.5 leads on the aggregate score, HiveSQL, and MySQL, the winners for the remaining engines are diverse: Claude Opus 4.8 achieves the best performance on PySpark ($83.8$), DeepSeek V4 Pro on PrestoSQL/Trino ($77.7$, tied with Gemini 3.5 Flash but at less than a third of its token cost), and DeepSeek V4 Flash on FlinkSQL ($85.0$). Several models exhibit pronounced variation across engines. For example, GLM 5.2 is competitive on PySpark and MySQL but performs poorly on HiveSQL. These results suggest that coverage across different engines, rather than the average score alone, provides a more informative measure of the capability of data engineering agents.

\noindent\textbf{Token cost does not track quality.}
Token consumption spans more than a $4\times$ range and shows no positive correlation with score. GPT 5.3 Codex is the most token-efficient ($271.4$k per task) yet ranks in the lower half, whereas Gemini 3.5 Flash consumes the most tokens by a wide margin ($973.8$k, up to $1529.0$k on PrestoSQL/Trino) while achieving only a mid-tier score. The best score-to-cost trade-offs come from models such as GPT 5.5 and Gemini 3.1 Pro, which combine top-tier accuracy with near-lowest token budgets, indicating that verbose exploration and repeated retries are often wasteful rather than beneficial.

\section{Analysis} \label{sec:analysis}
\subsection{Rule-Based Grading Is Irreplaceable}
A natural question is whether a generic LLM judge~\cite{zheng2023judging,li2024leveraging} could replace our case-specific deterministic grading scripts. To investigate this, we construct an LLM-judge prompt for each of the 100 benchmark cases, with each prompt faithfully reflecting the corresponding evaluation rubric, including its scoring dimensions, maximum scores, and weighting scheme. We then use two language models, GLM 5.2 and DeepSeek V4 Pro, to independently evaluate the same outputs produced by the Claude Opus 4.8 agent. Each model scores every output three times. The scores produced by our rule-based grading scripts are treated as the ground truth (GT), and all results are reported on a 0--100 scale.

\begin{table}[!htbp]
\centering
\caption{Comparison of LLM-as-Judge against Rule-Based (our) ground-truth scoring on the full 100-case DataClawEval suite, using the Claude Opus 4.8 agent outputs. Two judge models each score the same outputs three times; Rule-Based (our) is bit-exact. A per-engine breakdown and per-case scatter appear in Appendix~\ref{app:judge}.}
\label{tab:llm_judge_overall}
\renewcommand{\arraystretch}{1.0}%
\resizebox{\columnwidth}{!}{%
\begin{tabular}{lccc}
\toprule
\textbf{Metric} & \textbf{GLM 5.2} & \textbf{DeepSeek} & \textbf{Rule-Based} \\
\midrule
Overall score   & 82.1 & 84.4 & 74.3 \\
Product score  & 90.4   & 91.0   & 52.1   \\
Process score  & 76.0   & 75.0   & 22.2   \\
Overall $\sigma$ (stability) & 6.35 & 1.99 & 0 \\
\bottomrule
\end{tabular}%
}
\end{table}

Three systematic failures emerge. First, both judges inflate every metric (Table~\ref{tab:llm_judge_overall}): the process score more than triples from a ground-truth 22.2 to 75.0--76.0 and the product score rises from 52.1 to 90.4--91.0 because the judges reward visible attempts at data exploration or self-verification through surface-level pattern matching rather than code execution. Consequently, the vast majority of the 100 per-case scores fall above the $y=x$ diagonal (Figure~\ref{fig:judge_scatter}, Appendix~\ref{app:judge}). Second, they are non-deterministic (per-case overall-score standard deviation of 6.35 for GLM 5.2 and 1.99 for DeepSeek V4 Pro, worst case exceeding 55 points) and repeatedly breach hard rubric caps, e.g., awarding up to 291\% and 200\% of the maximum on a time-calculation and a field-matching dimension; such randomness lets an agent's leaderboard ranking shift on sampling noise alone. Third, the bias is strongly engine-dependent: HiveSQL is inflated most (a ground-truth score of 65.0 is rated 81.3--82.4), while FlinkSQL exhibits the largest bidirectional error. In particular, the judges severely penalize some runtime-correct streaming outputs while assigning high scores to others that fail execution-based checks, revealing their inability to reliably assess temporal streaming semantics. Detailed case-level contrasts are provided in Appendix~\ref{app:judge} and visualized by the widely dispersed FlinkSQL points in Figure~\ref{fig:judge_scatter}.

These failures are not incidental but structural consequences of text-only evaluation: rule-based grading executes each output against a live database and checks results row by row, whereas an LLM judge never observes runtime behavior. We therefore adopt rule-based grading as the definitive evaluation protocol for our benchmark.

\subsection{Model Stability Analysis}\label{sec:stability}

\begin{figure}[t]
    \centering
    \includegraphics[width=1.0\linewidth]{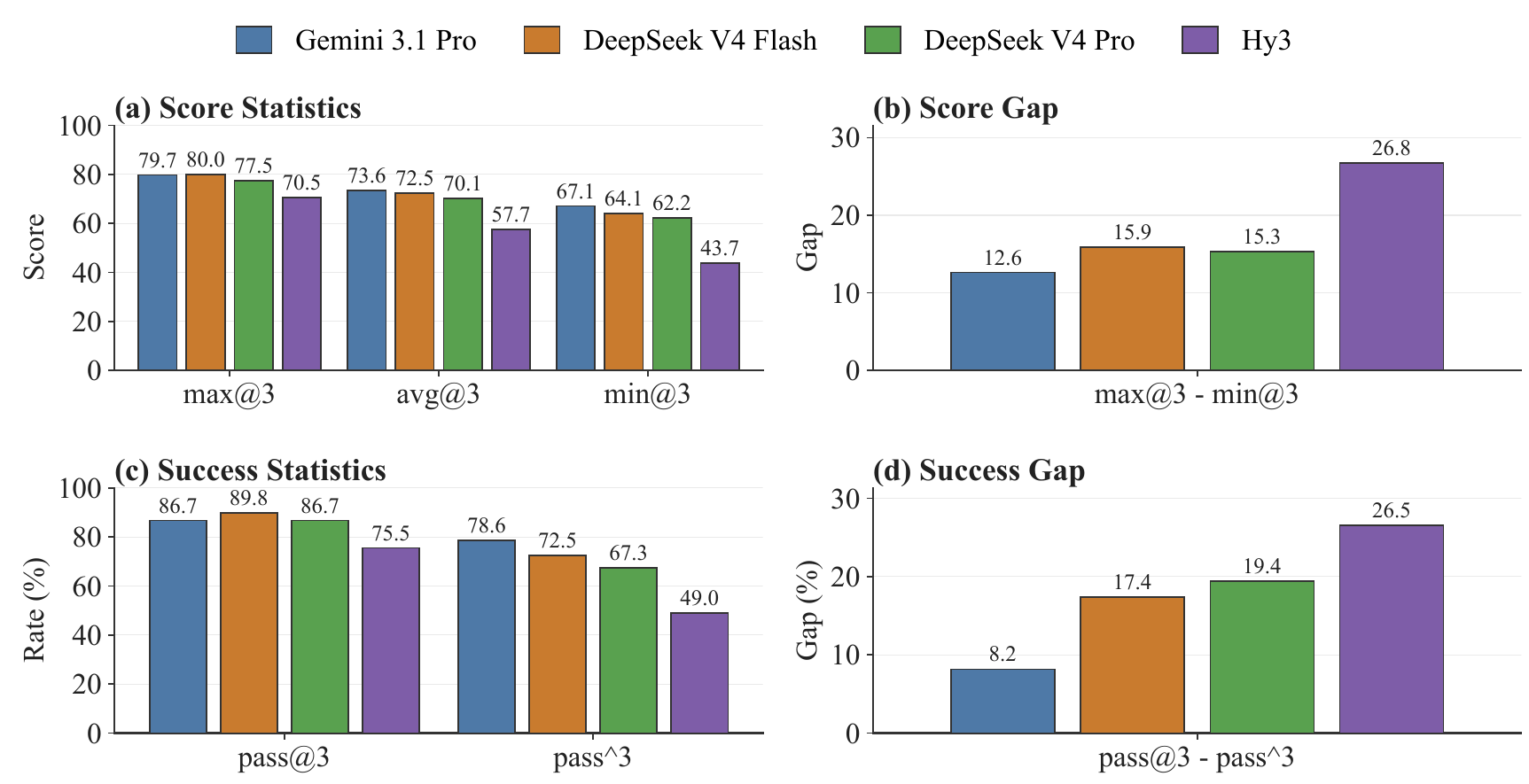}
    \caption{Run-to-run stability of four representative agents over three independent executions per task. \textbf{Top-left}: max@3, avg@3, and min@3 of the overall score. \textbf{Top-right}: the max@3$-$min@3 score gap. \textbf{Bottom-left}: pass@3 (solved in any run) and pass\textasciicircum{}3 (solved in all three runs). \textbf{Bottom-right}: the pass@3$-$pass\textasciicircum{}3 gap.}
    \label{fig:stability and Consistency}
\end{figure}

Figure~\ref{fig:stability and Consistency} evaluates model stability across three independent runs using both score-based (max@3, avg@3, min@3) and success-based (pass@3, pass\textasciicircum{}3) metrics. Gemini 3.1 Pro exhibits the most consistent behavior, with the smallest score variation (12.63 points) and the narrowest gap between pass@3 and pass\textasciicircum{}3 (8.16\%), indicating highly reproducible performance across repeated executions. DeepSeek V4 Flash and DeepSeek V4 Pro achieve competitive peak performance but exhibit noticeably larger run-to-run fluctuations, with score gaps of 15.92 and 15.31 points, respectively, and pass@3$-$pass\textasciicircum{}3 gaps of 17.35\% and 19.39\%. In contrast, Hy3 shows the weakest stability: although its best-case score reaches 70.49, its minimum score drops to 43.73, resulting in a 26.76-point gap, while its pass\textasciicircum{}3 decreases from 75.51\% to only 48.98\%.

Overall, these results demonstrate that high peak performance does not necessarily imply robust or reproducible execution. Models with similar max@3 scores can differ substantially in their worst-case performance and pass\textasciicircum{}3 results, suggesting that single-run evaluation may overestimate practical capability. Repeated-run evaluation therefore provides a more comprehensive assessment of data engineering agents by explicitly measuring execution robustness and consistency. This distinction is especially consequential in production data engineering, where a pipeline that succeeds only intermittently is effectively unreliable: an agent whose peak score masks a low min@3 or a fragile pass\textasciicircum{}3 cannot be trusted to run unattended on recurring workloads. We therefore report the full max@3/avg@3/min@3 and pass@3/pass\textasciicircum{}3 spectrum rather than any single aggregate, so that stability is treated as a first-class dimension of agent quality alongside peak accuracy.

\begin{figure}[t]
    \centering
    \includegraphics[width=1.0\linewidth]{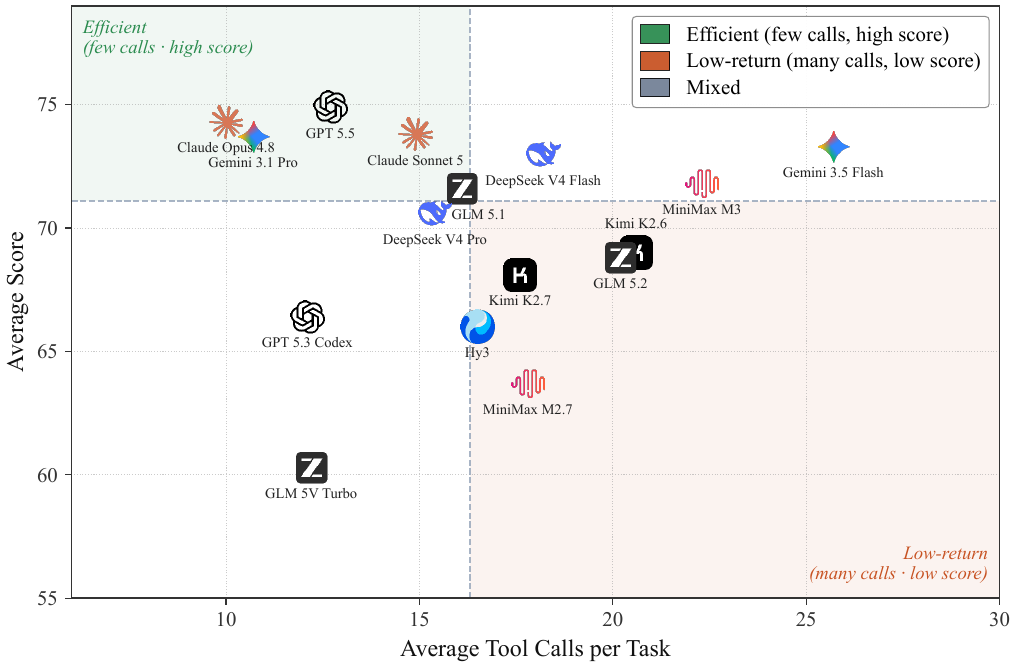}
    \caption{Average tool calls per task (x-axis) vs.\ average overall score (y-axis) for the 16 agents. Axes are split at their medians (calls $\approx 16.3$, score $\approx 71.1$): green marks efficient agents with few calls and high score, orange marks inefficient agents with many calls and low score, and grey marks the remaining agents.}
    \label{fig:tool call}
\end{figure}

\subsection{Tool Call Efficiency and Task Performance}

Figure~\ref{fig:tool call} relates average tool calls per task to average overall score for the 16 agents. Tool-call frequency and score are nearly uncorrelated; more calls do not translate into higher scores.

The strongest models cluster in the green efficient zone (upper-left): GPT 5.5, Claude Opus 4.8, Claude Sonnet 5, Gemini 3.1 Pro, and GLM 5.1 achieve top-tier scores with below-median call counts. In contrast, the orange inefficient zone (lower-right) groups the most frequent callers that nonetheless score below the median, including Kimi K2.6/K2.7, GLM 5.2, MiniMax M2.7, and Hy3.

These results indicate that the differentiator between leading and lagging agents is tool-use efficiency rather than call volume. Agents that substitute trial-and-error for precise planning tend to accumulate calls without improving their final artifacts.

\section{Related Work} \label{sec:related-work}

\noindent\textbf{Single-Step Text-to-SQL.}
Text-to-SQL is a long-standing testbed for semantic parsing, from WikiSQL~\cite{zhong2017wikisql} and cross-domain Spider~\cite{yu2018spider} to BIRD~\cite{li2023bird} on large, dirty databases and Spider 2.0~\cite{lei2025spider2} on enterprise cloud warehouses. These benchmarks still center on synthesizing a SQL answer from a fixed schema, rather than the full engineering workflow of schema exploration, multi-engine implementation, iterative debugging, and artifact materialization.

\noindent\textbf{Data Agent Benchmarks.}
Data-specific agents have been evaluated along three lines: code-execution data analysis benchmarks run Python/SQL/Bash to reach an answer (InfiAgent-DABench~\cite{hu2024infiagent}, DA-Code~\cite{huang2024dacode}, DABstep~\cite{egg2025dabstep}); heterogeneous and exploratory analysis benchmarks reason over mixed sources (FDABench~\cite{wang2026fdabench}, DataClawBench~\cite{zhang2025dataclawbench}, FinanceBench~\cite{islam2023financebench}); and ELT-Bench~\cite{jin2025eltbench} takes a first step toward engineering via pipeline construction on a single dbt/Airbyte stack. DataClawEval likewise targets data agents, but is the first to cover enterprise data engineering across five batch and streaming engines under a unified, deterministic, rule-based protocol that grades both artifacts and process.

\noindent\textbf{General Agent and Code Benchmarks.}
General-purpose agent and code-generation benchmarks (e.g., AgentBench~\cite{liu2024agentbench}, ToolLLM~\cite{qin2024toolllm}, MINT~\cite{wang2024mint}, DS-1000~\cite{lai2023ds1000}, WildClawBench~\cite{ding2026wildclawbench}) measure tool use, multi-turn interaction, and program synthesis, but are not tailored to data workflows and omit database semantics, multi-engine execution, and pipeline construction.

\noindent\textbf{Evaluation Paradigms.}
Execution-grounded protocols (e.g., execution accuracy~\cite{li2023bird,lei2025spider2} and unit tests~\cite{chen2021evaluating,lai2023ds1000}) are deterministic but usually match only a single final answer, whereas LLM-as-a-Judge~\cite{zheng2023judging,li2024leveraging} scales to open-ended outputs yet suffers score inflation, bias, and unstable verdicts. As Section~\ref{sec:analysis} shows, LLM judges systematically overrate DataClawEval and breach rubric caps, so we adopt case-specific, deterministic, rule-based scripts that execute each output against live databases and grade both product and process.

\section{Conclusion} \label{sec:conclusion}

We introduced DataClawEval, the first benchmark designed to evaluate whether autonomous agents can complete real industrial data engineering workflows instead of merely generating code for them. Grounded in production-grade implementations and reconstructed through a human-in-the-loop pipeline with differential validation, DataClawEval contains 100 executable tasks spanning batch and streaming workloads across PySpark, HiveSQL, MySQL, PrestoSQL/Trino, and FlinkSQL. Within a unified industrial agent harness, agents must inspect data, implement and debug programs, and materialize correct artifacts in isolated environments, which are assessed by case-specific deterministic graders. Our evaluation of 16 frontier agents shows that this capability remains far from solved: the best overall score is only $74.9$, no agent dominates across engines, and strengths remain highly engine-specific, while repeated runs and inefficient tool use expose practical reliability gaps. The large errors and instability of LLM judges further demonstrate that execution-grounded, rule-based evaluation is indispensable. By releasing the tasks, environments, and graders, DataClawEval establishes a reproducible testbed for measuring and advancing end-to-end data engineering agents.

%%
%% Bibliography
\bibliographystyle{ACM-Reference-Format}
\bibliography{sample-base}

\clearpage
\appendix

\onecolumn
\section{Task Listing}
\label{sec:task-listing}

\begin{multicols}{2}
Table~\ref{tab:task-listing} lists all $100$ tasks in DataClawEval with their execution engine and business domain.
The five engines (PySpark, MySQL, HiveSQL, PrestoSQL/Trino, FlinkSQL) are shown as grouped rows, and each task falls into one of five business domains:

\textbf{Ops \& Resource Governance} (30 tasks), covering GPU/notebook instance governance, message-queue and quota management, resource-cost aggregation, and inference-service capacity planning;
\textbf{Data Analytics \& User Growth} (30 tasks), covering user activity and retention, PV/UV statistics, real-time windowed metrics, and funnel/TopN analysis;
\textbf{Security \& Risk Control} (16 tasks), covering security scanning, threat-intelligence correlation, anomaly detection, and risk scoring;
\textbf{Content, Community \& Dev-Efficiency} (14 tasks), covering content quality scoring, community wide-table modeling, and code-review/engineering-efficiency analytics;
and \textbf{Advertising \& Marketing} (10 tasks), covering ad attribution, campaign tagging, and marketing-effect aggregation.
\end{multicols}

\setlength{\LTcapwidth}{\textwidth}
\small
\begin{longtable}{@{} l l p{7.2cm} l @{}}
\caption{Per-task listing for all 100 DataClawEval tasks, grouped by execution engine and annotated with business domain.}
\label{tab:task-listing} \\
\toprule
\textbf{Engine} & \textbf{Task ID} & \textbf{Task Name} & \textbf{Business Domain} \\
\midrule
\endfirsthead
\caption[]{Per-task listing for all 100 DataClawEval tasks (continued).} \\
\toprule
\textbf{Engine} & \textbf{Task ID} & \textbf{Task Name} & \textbf{Business Domain} \\
\midrule
\endhead
\midrule
\multicolumn{4}{r}{\textit{Continued on next page}} \\
\endfoot
\bottomrule
\endlastfoot
\multirow{20}{*}{PySpark} & pyspark\_001 & Passive CS Satisfaction Attribution Comparison & Data Analytics \& User Growth \\
 & pyspark\_002 & Code Review Note Diff ODS-to-DWV Sync & Content, Community \& Dev-Eff. \\
 & pyspark\_003\_en & Code Review Diff ODS-to-DWV Sync & Content, Community \& Dev-Eff. \\
 & pyspark\_004 & Sales Lead Enterprise Email Brand Detail & Advertising \& Marketing \\
 & pyspark\_005\_en & AI Search Parse Quality Daily (v2) & Security \& Risk Control \\
 & pyspark\_006\_en & AI Search Parse Quality Daily & Security \& Risk Control \\
 & pyspark\_007\_en & Cloud-Check User Active Status Sync & Security \& Risk Control \\
 & pyspark\_008\_en & API Cloud-Check Log Cleansing & Security \& Risk Control \\
 & pyspark\_009\_en & Anti-Spam H5 Work Order Log Sync & Security \& Risk Control \\
 & pyspark\_010 & PM Workitem Changes Deduplication & Content, Community \& Dev-Eff. \\
 & pyspark\_011\_en & Phone Prefix Quality Score Statistics & Security \& Risk Control \\
 & pyspark\_012 & Web Parse Top500 Quality Stats & Security \& Risk Control \\
 & pyspark\_013 & Bank URL Safety Relationship Filter & Security \& Risk Control \\
 & pyspark\_014 & Experiment Plan History Query & Data Analytics \& User Growth \\
 & pyspark\_015 & User Tag Sync ETL & Data Analytics \& User Growth \\
 & pyspark\_016 & Code Review Organization Ranking & Content, Community \& Dev-Eff. \\
 & pyspark\_017\_en & Advertising Metric Anomaly Detection & Data Analytics \& User Growth \\
 & pyspark\_018\_en & SOA Dependency Analysis & Ops \& Resource Governance \\
 & pyspark\_019\_en & Campaign User Label Computation & Advertising \& Marketing \\
 & pyspark\_020 & Mini-Program H5 Monitoring & Data Analytics \& User Growth \\
\multirow{20}{*}{MySQL} & mysql\_001 & Real-Time Ad RPM Data Filter \& Rename & Advertising \& Marketing \\
 & mysql\_002\_en & Inference Service Replica Forecast & Ops \& Resource Governance \\
 & mysql\_003 & Offline Inference Task Feature Wide Table & Ops \& Resource Governance \\
 & mysql\_004\_en & Ceph Path Coldness Score Statistics & Ops \& Resource Governance \\
 & mysql\_005\_en & GPU Inference Platform P90 Traffic Forecast & Ops \& Resource Governance \\
 & mysql\_006 & Inference Service Traffic \& Resource Forecast & Ops \& Resource Governance \\
 & mysql\_007 & Hot Table Governance Metadata Wide Table & Ops \& Resource Governance \\
 & mysql\_008 & Consumer Group Governance Detail Extraction & Ops \& Resource Governance \\
 & mysql\_009\_en & Databus Consumer Group Cost Aggregation & Ops \& Resource Governance \\
 & mysql\_010\_en & Table Heat Field \& Query User Stats & Ops \& Resource Governance \\
 & mysql\_011\_en & Low-Value Task Benefit Aggregation & Ops \& Resource Governance \\
 & mysql\_012 & Notebook Pipeline GPU Utilization Stats & Ops \& Resource Governance \\
 & mysql\_013\_en & Notebook Cross-Day Instance Pod Detail & Ops \& Resource Governance \\
 & mysql\_014 & Notebook Instance Detail by Day \& Engine & Ops \& Resource Governance \\
 & mysql\_015\_en & Exposure Log First Exposure Time & Advertising \& Marketing \\
 & mysql\_016 & App X-Model Org Relation Hour Migration & Data Analytics \& User Growth \\
 & mysql\_017 & News Exposure Data Full Migration & Content, Community \& Dev-Eff. \\
 & mysql\_018\_en & URL Safety Detection Access Statistics & Security \& Risk Control \\
 & mysql\_019 & Sensor Event Wide Table Join & Content, Community \& Dev-Eff. \\
 & mysql\_020\_en & Volume-Lift Candidate Ad List & Advertising \& Marketing \\
\multirow{28}{*}{HiveSQL} & hivesql\_001\_en & Message Queue Topic Dimension Table Sync & Ops \& Resource Governance \\
 & hivesql\_002 & Notebook Ray Cross-Day Instance Detail & Ops \& Resource Governance \\
 & hivesql\_003 & Notebook Killed Ray Instance Statistics & Ops \& Resource Governance \\
 & hivesql\_004 & App Group Product Info Daily Snapshot & Ops \& Resource Governance \\
 & hivesql\_005 & MQ Topic Production Feature Governance & Ops \& Resource Governance \\
 & hivesql\_006\_en & Databus Topic Access Cost Aggregation & Ops \& Resource Governance \\
 & hivesql\_007\_en & Notebook Instance Runtime Statistics & Ops \& Resource Governance \\
 & hivesql\_008\_en & Shuffle Split Tuning Rule Hit Count & Ops \& Resource Governance \\
 & hivesql\_009\_en & Task Instance GPU Card-Hour Statistics & Ops \& Resource Governance \\
 & hivesql\_010 & Killed Notebook Instance Pod Runtime Detail & Ops \& Resource Governance \\
 & hivesql\_011 & Killed Notebook Instance Detail & Ops \& Resource Governance \\
 & hivesql\_012\_en & Social Book Tag Incremental Filter & Content, Community \& Dev-Eff. \\
 & hivesql\_013 & Ad Material Action Metrics Summary & Advertising \& Marketing \\
 & hivesql\_014\_en & Add-Friend Behavior Cluster Aggregation & Security \& Risk Control \\
 & hivesql\_015\_en & User Tag Preset Incremental Insert & Data Analytics \& User Growth \\
 & hivesql\_016 & Daily Active Users by Channel & Data Analytics \& User Growth \\
 & hivesql\_017 & SDK Hourly Active Users Statistics & Data Analytics \& User Growth \\
 & hivesql\_018\_en & First-Level Comment Score Computation & Content, Community \& Dev-Eff. \\
 & hivesql\_019\_en & Potential Risk Domain Extraction & Security \& Risk Control \\
 & hivesql\_020 & Dictionary Item Hour Monitoring Stats & Ops \& Resource Governance \\
 & hivesql\_021\_en & Chinese Add-Friend Content Filter & Security \& Risk Control \\
 & hivesql\_022 & Audit Data Parse and Transcode & Content, Community \& Dev-Eff. \\
 & hivesql\_023\_en & User Open-Start-Path CUBE Aggregation & Data Analytics \& User Growth \\
 & hivesql\_024 & Relationship Strength Score Computation & Data Analytics \& User Growth \\
 & hivesql\_025 & Product Data Source Quality Scoring & Content, Community \& Dev-Eff. \\
 & hivesql\_026\_en & Multi-Scene Business Detail Merge & Advertising \& Marketing \\
 & hivesql\_027 & Digital Tool Platform Daily Report & Content, Community \& Dev-Eff. \\
 & hivesql\_028\_en & Feed Aggregator Check Result Explode & Content, Community \& Dev-Eff. \\
\multirow{12}{*}{\makecell{PrestoSQL/\\Trino}} & prestosql\_001 & Security Scan Instance Detail Analysis & Security \& Risk Control \\
 & prestosql\_002\_en & Ad ASA Attribution Migration & Advertising \& Marketing \\
 & prestosql\_003 & IM Group Offline Report Sub-Partition Filter & Security \& Risk Control \\
 & prestosql\_004 & Novel UDS Algorithm Book Relation Filter & Content, Community \& Dev-Eff. \\
 & prestosql\_005\_en & Hot Table Governance Metadata Aggregation & Ops \& Resource Governance \\
 & prestosql\_006\_en & News Plugin Send Process Aggregation & Content, Community \& Dev-Eff. \\
 & prestosql\_007\_en & APK Threat Scan GPU Card-Hour Stats & Ops \& Resource Governance \\
 & prestosql\_008\_en & Report WUID Detail Filtering & Advertising \& Marketing \\
 & prestosql\_009 & Data Pipeline Killed Instance Analysis & Ops \& Resource Governance \\
 & prestosql\_010 & IOC MTTD Metric Computation & Security \& Risk Control \\
 & prestosql\_011 & Sensitive Interface Union \& Aggregation & Security \& Risk Control \\
 & prestosql\_012\_en & Data Quality Check Pipeline GPU Analysis & Ops \& Resource Governance \\
\multirow{20}{*}{FlinkSQL} & flinksql\_001\_en & Order Stream and Payment Stream Interval Join & Data Analytics \& User Growth \\
 & flinksql\_002\_en & Order-Payment Interval Join + Windowed Agg. & Data Analytics \& User Growth \\
 & flinksql\_003\_en & Word Frequency Tumbling Window Statistics & Data Analytics \& User Growth \\
 & flinksql\_004\_en & Dual Window Aggregation (Tumble + Hop) & Data Analytics \& User Growth \\
 & flinksql\_005 & Tumbling Window TopN Product Ranking & Data Analytics \& User Growth \\
 & flinksql\_006 & Anomaly Event Filter + Window TopN & Data Analytics \& User Growth \\
 & flinksql\_007 & User Behavior Tumbling Window Statistics & Data Analytics \& User Growth \\
 & flinksql\_008\_en & Session Window User Behavior Aggregation & Data Analytics \& User Growth \\
 & flinksql\_009\_en & Tumbling Window PV/UV Statistics & Data Analytics \& User Growth \\
 & flinksql\_010 & Sliding Window (HOP) Shop Statistics & Data Analytics \& User Growth \\
 & flinksql\_011\_en & Tumbling Window Product TopN Ranking & Data Analytics \& User Growth \\
 & flinksql\_012\_en & Session Window User Behavior Statistics & Data Analytics \& User Growth \\
 & flinksql\_013 & Cumulative Window Daily Active UV & Data Analytics \& User Growth \\
 & flinksql\_014 & Interval Join + Window Aggregation & Data Analytics \& User Growth \\
 & flinksql\_015 & Regular Join Impression-Click Association & Advertising \& Marketing \\
 & flinksql\_016 & Tumbling Window Multi-Dim Agg. (GROUPING SETS) & Data Analytics \& User Growth \\
 & flinksql\_017\_en & Row-Level TopN (Windowless) & Data Analytics \& User Growth \\
 & flinksql\_018\_en & Dual Output (Detail + Windowed Agg.) & Data Analytics \& User Growth \\
 & flinksql\_019 & Global Group-By + Filter Statistics & Data Analytics \& User Growth \\
 & flinksql\_020 & Window Deduplication (TVF Dedup) & Data Analytics \& User Growth \\
\end{longtable}

\twocolumn[
\begin{@twocolumnfalse}
\section{Analysis of Low-Score Runs}
\label{sec:failure-mode}
\centering
\small
\setlength{\tabcolsep}{5pt}
\captionof{table}{Per-engine low-score rates and the taxonomy of execution errors raised by low-score runs. ``Low-score'' and ``Low\%'' are the number and share of an engine's runs that score below $50$ on the 0--100 scale. Error categories are multi-label---a low-score run may raise several, or none---and are counted over that engine's low-score runs.}
\label{tab:failure-summary}
\begin{tabular*}{\textwidth}{@{\extracolsep{\fill}} l rrr rrrrrr @{}}
\toprule
 & \multicolumn{3}{c}{} & \multicolumn{6}{c}{Execution-error incidences (multi-label)} \\
\cmidrule(lr){5-10}
\textbf{Engine} & Runs & Low-score & Low\% & Missing & Syntax & Type & Runtime & Env & Resource \\
\midrule
PySpark          & 320  & 34  & 10.6 & 5  & 1  & 4  & 19 & 12 & 1  \\
HiveSQL          & 448  & 117 & 26.1 & 5  & 5  & 29 & 11 & 8  & 4  \\
MySQL            & 320  & 31  & 9.7  & 11 & 0  & 1  & 12 & 0  & 0  \\
PrestoSQL/Trino  & 192  & 49  & 25.5 & 16 & 13 & 1  & 4  & 22 & 12 \\
FlinkSQL         & 320  & 61  & 19.1 & 2  & 15 & 12 & 3  & 4  & 36 \\
\midrule
\textbf{All}     & \textbf{1600} & \textbf{292} & \textbf{18.2} & \textbf{39} & \textbf{34} & \textbf{47} & \textbf{49} & \textbf{46} & \textbf{53} \\
\bottomrule
\end{tabular*}
\vspace{8pt}
\end{@twocolumnfalse}
]

We examine the $292$ runs whose rule-based score falls below $50$ on the 0--100 scale, which is $18.2\%$ of the $1{,}600$ total. Low scores are not spread evenly; they concentrate in a few tasks and a few engines. Table~\ref{tab:failure-summary} reports the per-engine low-score rates and error taxonomy.

\begin{figure}[t]
\centering
\includegraphics[width=\columnwidth]{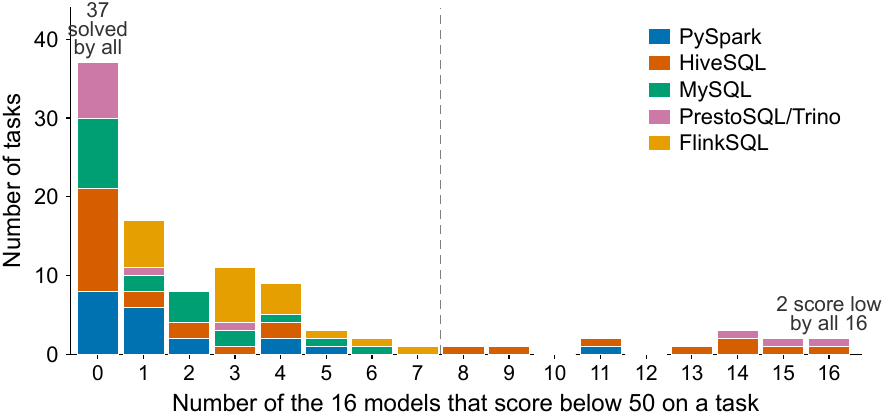}
\caption{Distribution of low scores across the $100$ DataClawEval tasks. The $x$-axis is the number of the $16$ models that score below $50$ on the 0--100 scale for a task; the $y$-axis counts tasks, stacked by execution engine. Most tasks are rarely scored low: $37$ receive no low scores at all, while about a dozen hard tasks, mostly in HiveSQL and PrestoSQL/Trino, account for roughly half of all low-score runs.}
\label{fig:failure}
\end{figure}

Across tasks (Figure~\ref{fig:failure}), $37$ of the $100$ receive no low scores at all, while about a dozen hard tasks account for roughly half of the low-score runs. By engine (Table~\ref{tab:failure-summary}), HiveSQL and PrestoSQL/Trino score low most often ($26.1\%$ and $25.5\%$). Every model is affected to some degree, with per-model low-score counts ranging from $11$ to $26$ out of $100$.

Low-score runs fail in two broad ways. About a third ($102/292$) are \emph{silent}: the program finishes and writes an output table, but the result is wrong. The rest raise at least one execution error at run time.

\section{Characterizing Timed-Out Runs}
\label{sec:timeout-confound}

Among the low-score runs, $79$ of the $1{,}600$ ($4.9\%$) reached the task's wall-clock limit before producing the required artifact (e.g., \texttt{result.sql}) and were scored $0$. Figure~\ref{fig:timeout-confound}(a) shows where these runs fall within each model's score distribution. For every model except GPT 5.5 they appear as a separate cluster of points at $0$, detached from the box that holds the bulk of that model's scores, and their number varies widely, from none to $22$ out of $100$ runs. GLM 5.2 makes the pattern most visible: its box lies at the top of the plot, yet it also carries the largest cluster of zeros.

Each timed-out run contributes a $0$, so it lowers a model's mean in proportion to how often timeouts occur. Figure~\ref{fig:timeout-confound}(b) quantifies this as the gap between each model's reported mean and its mean over the remaining runs. The gap grows with the timeout share but stays within $2.8$ points for $12$ of the $16$ models, leaving their ranking unchanged. GLM 5.2 is again the exception, losing $13.3$ points to a $22\%$ timeout rate. Its reported mean of $68.8$ therefore ranks eleventh, whereas its mean over completed runs is $82.1$, the highest of all $16$ models. This shows that a high timeout rate can substantially depress a capable model's reported score.

\begin{strip}
\centering
\includegraphics[width=\textwidth]{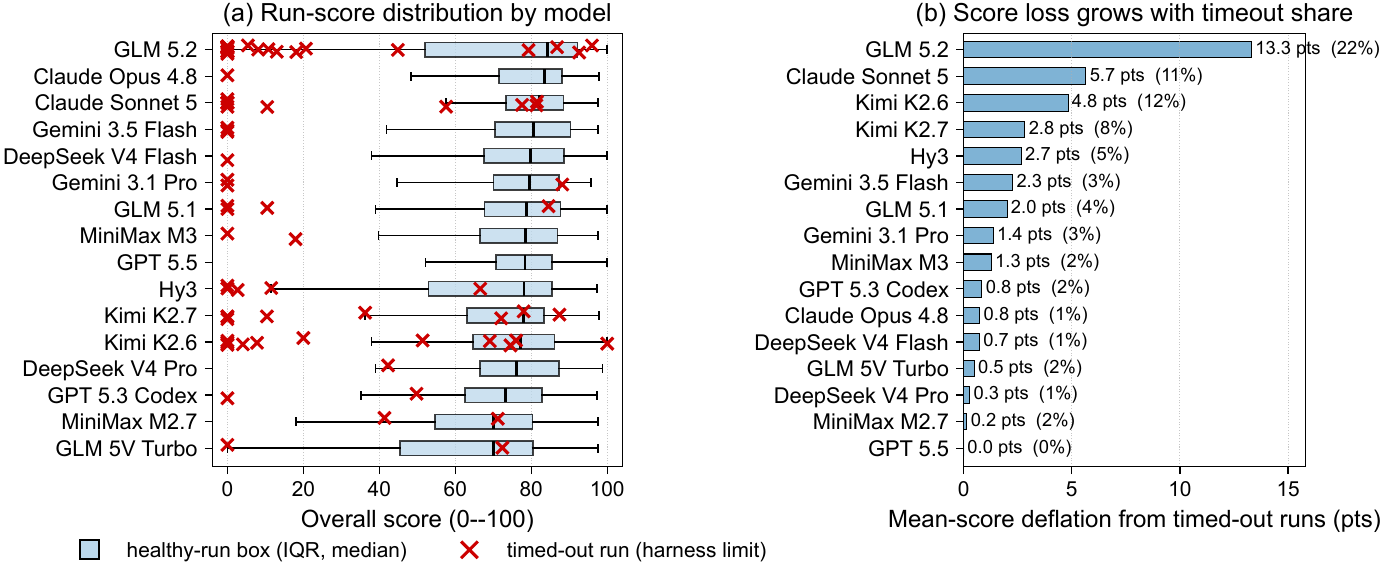}
\captionof{figure}{Characterization of the $79$ runs terminated at the harness wall-clock limit. \textbf{(a)}~Per-model distribution of the $100$ run scores (box: healthy-run IQR and median; red $\times$: timed-out runs). Timed-out runs affect every model except GPT 5.5, with incidence spanning $0$--$22\%$. \textbf{(b)}~Mean-score deflation attributable to timed-out runs (mean over healthy runs minus the reported mean), sorted; bar-end labels additionally show each model's timeout share. For $12$ of $16$ models the deflation is at most $2.8$ points; only one model is materially affected.}
\label{fig:timeout-confound}
\end{strip}

\section{Bilingual Performance Analysis}
\label{sec:bilingual}

Task prompts in DataClawEval come in two languages: $50$ are in English and
$50$ in Chinese, following the language of the original production logs. The
split is balanced within each engine ($10$/$10$ for PySpark, MySQL, and
FlinkSQL, $6$/$6$ for PrestoSQL/Trino, and $14$/$14$ for HiveSQL), giving $800$
English and $800$ Chinese runs across the $16$ agents. We recompute the average
run-level overall score (0--100 scale) on each subset;
Table~\ref{tab:bilingual-model} gives the per-model breakdown and
Table~\ref{tab:bilingual-engine} the per-engine one.

Averaged over all agents, the English subset scores slightly higher than the
Chinese subset ($71.7$ vs.\ $68.2$), and $14$ of the $16$ agents show a positive
English-subset difference. These are disjoint task subsets, not parallel
translations, so the differences are confounded by task difficulty and cannot
be attributed to prompt language alone. The gaps are typically a few points,
comparable to the run-to-run variance in Section~\ref{sec:stability}, and a few
models (e.g., GLM 5V Turbo) show a small negative difference.

\begin{table}[t]
\centering
\small
\caption{Bilingual performance of the $16$ agents on DataClawEval. \emph{Overall} is the mean overall score across all $100$ tasks; \emph{English} and \emph{Chinese} are the means over the $50$ English and $50$ Chinese tasks, respectively; \emph{EN$-$ZH} is their English-subset difference (a positive value indicates a higher score on the English subset). All scores are on a 0--100 scale, and models are ordered by overall score.}
\label{tab:bilingual-model}
\begin{tabular}{lcccc}
\toprule
\textbf{Model} & \textbf{Overall} & \textbf{English} & \textbf{Chinese} & \textbf{EN$-$ZH} \\
\midrule
GPT 5.5            & 74.9 & 76.9 & 72.9 & $+4.0$ \\
Claude Opus 4.8    & 74.3 & 75.1 & 73.6 & $+1.5$ \\
Claude Sonnet 5    & 73.8 & 73.5 & 74.1 & $-0.6$ \\
Gemini 3.1 Pro     & 73.7 & 75.3 & 72.1 & $+3.2$ \\
Gemini 3.5 Flash   & 73.3 & 74.3 & 72.3 & $+2.0$ \\
DeepSeek V4 Flash  & 73.0 & 76.2 & 69.7 & $+6.5$ \\
MiniMax M3         & 71.8 & 74.3 & 69.4 & $+4.9$ \\
GLM 5.1            & 71.6 & 76.5 & 66.6 & $+9.9$ \\
DeepSeek V4 Pro    & 70.6 & 74.3 & 66.9 & $+7.4$ \\
Kimi K2.6          & 69.0 & 69.6 & 68.3 & $+1.3$ \\
GLM 5.2            & 68.8 & 71.8 & 65.8 & $+6.0$ \\
Kimi K2.7          & 68.1 & 70.3 & 65.9 & $+4.4$ \\
GPT 5.3 Codex      & 66.4 & 68.1 & 64.7 & $+3.4$ \\
Hy3                & 66.0 & 66.8 & 65.1 & $+1.7$ \\
MiniMax M2.7       & 63.7 & 66.2 & 61.2 & $+5.0$ \\
GLM 5V Turbo       & 60.3 & 58.0 & 62.5 & $-4.5$ \\
\midrule
\textbf{Mean}      & \textbf{70.0} & \textbf{71.7} & \textbf{68.2} & $\mathbf{+3.5}$ \\
\bottomrule
\end{tabular}
\end{table}

\begin{table}[t]
\centering
\small
\caption{Bilingual performance broken down by execution engine, averaged over all $16$ agents. Scores are on a 0--100 scale.}
\label{tab:bilingual-engine}
\begin{tabular}{lcccc}
\toprule
\textbf{Engine} & \textbf{Overall} & \textbf{English} & \textbf{Chinese} & \textbf{EN$-$ZH} \\
\midrule
PySpark          & 69.1 & 67.9 & 70.3 & $-2.4$ \\
HiveSQL          & 61.7 & 65.9 & 57.5 & $+8.4$ \\
MySQL            & 81.4 & 83.9 & 79.0 & $+4.9$ \\
PrestoSQL/Trino  & 72.2 & 74.3 & 70.2 & $+4.1$ \\
FlinkSQL         & 69.5 & 70.0 & 69.1 & $+0.9$ \\
\midrule
\textbf{All}     & \textbf{70.0} & \textbf{71.7} & \textbf{68.2} & $\mathbf{+3.5}$ \\
\bottomrule
\end{tabular}
\end{table}

\section{LLM-as-Judge Evaluation Details}
\label{app:judge}

\begin{figure}[t]
    \centering
    \includegraphics[width=\linewidth]{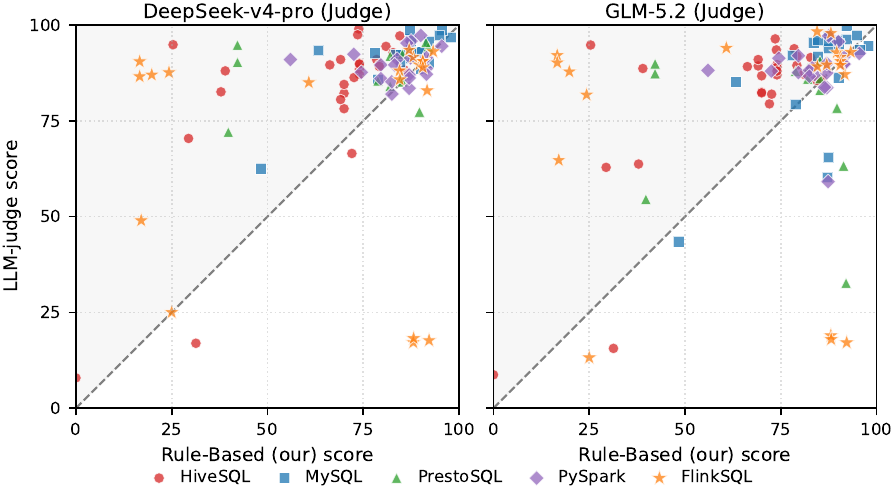}
    \caption{Per-case LLM-judge score versus Rule-Based (our) on the 100-case suite (Claude Opus 4.8 outputs), for DeepSeek V4 Pro (left) and GLM 5.2 (right). Each point is one case; the dashed line is $y=x$ and the shaded region marks overestimation. Most points lie above the diagonal (systematic inflation), while FlinkSQL cases (stars) scatter to both extremes.}
    \label{fig:judge_scatter}
\end{figure}

This appendix expands the LLM-as-Judge study of Section~\ref{sec:analysis}. Under the same protocol---each of the 100 cases scored three times by GLM 5.2 and DeepSeek V4 Pro against the Rule-Based (our) ground truth---Figure~\ref{fig:judge_scatter} plots every per-case judge score against its ground-truth value, and Table~\ref{tab:judge_module} reports the breakdown by engine.

\begin{table}[t]
\centering
\small
\caption{Per-engine overall score (0--100), averaged over the three judge runs, versus the Rule-Based (our) ground truth.}
\label{tab:judge_module}
\setlength{\tabcolsep}{5pt}
\begin{tabular}{lcccc}
\toprule
\textbf{Engine} & \textbf{\#} & \textbf{GLM 5.2} & \textbf{DeepSeek V4 Pro} & \shortstack{\textbf{Rule-Based}\\\textbf{(our)}} \\
\midrule
MySQL     & 20 & 87.7 & 91.8 & 85.1 \\
PySpark   & 20 & 87.9 & 89.6 & 83.8 \\
PrestoSQL & 12 & 77.4 & 87.5 & 74.7 \\
HiveSQL   & 28 & 81.3 & 82.4 & 65.0 \\
FlinkSQL  & 20 & 74.9 & 72.9 & 67.0 \\
\bottomrule
\end{tabular}
\end{table}

\noindent\textbf{Overestimation is dimension-structural.} The inflation concentrates in dimensions that require executing the code to verify. On the process side, self-verification ($I$) reaches a ground-truth rate of only 42.4\% but is rated 80.6--82.5\% by the judges, and exploration ($G$) rises from 46.6\% to 58.0--62.9\%---the judges credit a described attempt without confirming it produced correct results. Several product dimensions also breach their hard caps: the time-calculation dimension (max 5) is awarded 13.3--14.5 (267--291\% of maximum), field-value matching (max 10) is awarded 18.5--20.0 (185--200\%), and a resource-join dimension (max 5, ground truth 0) is awarded 6.9--7.8. Conversely, dimensions whose correctness is invisible in text are \emph{under}-scored, e.g., shuffle logic (max 20, ground truth 20.0) receives only 3.7--6.0.

\noindent\textbf{Non-reproducibility.} Averaged over the suite, the per-case standard deviation of the overall score is 6.35 for GLM 5.2 versus 1.99 for DeepSeek V4 Pro. GLM 5.2 is the more erratic: its worst single-case deviations exceed 55 points (e.g., \texttt{mysql\_020} at $\sigma{=}0.557$ and \texttt{prestosql\_008} at $\sigma{=}0.547$ on the 0--1 scale), so a case can flip several grade bands across runs. Because the rule-based grader is bit-exact ($\sigma{=}0$), such randomness alone can reorder an agent's leaderboard position.

\noindent\textbf{Streaming semantics defeat the judges.} FlinkSQL drives the largest bidirectional error. Streaming-correct outputs are heavily penalized---\texttt{flinksql\_008} (ground truth 0.92) is rated 0.17, and \texttt{flinksql\_011}/\texttt{012} (ground truth 0.88) are rated 0.17--0.19---while streaming-incorrect ones are inflated---\texttt{flinksql\_005} and \texttt{flinksql\_009} (ground truth 0.17) are rated 0.90--0.92, and \texttt{flinksql\_007} (ground truth 0.24) is rated 0.82--0.88. Since the judges read only the query text and never observe windowed, event-time execution, these errors cancel in the engine average (a modest 72.9--74.9 against a ground-truth 67.0) yet corrupt every per-case ranking, confirming that runtime execution is indispensable for evaluating streaming tasks.

\section{Representative Task Case Studies}
\label{sec:case-studies}
\begingroup
\small
\setlength{\parskip}{0pt}

To illustrate the diversity and complexity of DataClawEval, we present one representative case from each of the five business domains. Each case includes the task prompt (condensed), key input tables, core processing logic, output requirements, and grading criteria with the artifact weight $\alpha$. The five cases span all five execution engines.

\subsection{Inference Service Replica Forecast}
\textit{Domain: Ops \& Resource Governance}
\label{sec:case-ops}

\textit{Task ID:} \texttt{mysql\_002\_en}\hfill\textit{Engine:} MySQL

\paragraph{Prompt.} For inference services with autoscaling (HPA/AHPA), expand hourly expected pod counts to 10-minute intervals, clip them using configured replica bounds $\times$ machine count, compute a bias correction coefficient from the past 24\,h of actual pod counts, and label the AHPA recommended status. Write results for partition \texttt{dt='2026050700'}.

\paragraph{Input Tables.} HPA configuration, AHPA recommendations, service information, hourly expected pod counts, and actual pod count features (five tables).

\paragraph{Processing Logic.}
\begin{enumerate}\setlength{\itemsep}{0pt}\setlength{\parsep}{0pt}
\item Filter: \texttt{status='done'}, not deleted, HPA/AHPA enabled; latest config.
\item Cross-join hourly expected with six 10-min offsets (0--50\,min).
\item Pod bounds = \texttt{min/max\_replicas} $\times$ \texttt{host\_num}.
\item Clip all eight expected fields: \texttt{GREATEST(LEAST(v, upper), lower)}.
\item Bias coefficient = actual $/$ clipped \texttt{expected\_pod\_avg\_qpm\_p90} (1.0 if 0).
\item AHPA status: no records $\to$ \texttt{no\_request}; all non-200 $\to$ \texttt{degraded}; else \texttt{normal}.
\end{enumerate}

\paragraph{Output.} 29-column table via \codeinline{INSERT INTO \dots\ SELECT}; 10-min \codeinline{agg\_time}; partition \texttt{dt='2026050700'}; MySQL InnoDB.

\paragraph{Grading} ($\alpha{=}0.7$): Executability 15, Schema 15, Row alignment 20, Numerical accuracy 40, Label correctness 10. Process 30\%: exploration 35, efficiency 40, self-verification 25.

\subsection{Order-Payment Interval Join + Windowed Aggregation}
\textit{Domain: Data Analytics \& User Growth}
\label{sec:case-analytics}

\textit{Task ID:} \texttt{flinksql\_002\_en}\hfill\textit{Engine:} FlinkSQL

\paragraph{Prompt.} Using two Flink \texttt{datagen} tables simulating an order stream and a payment stream (1{,}000 rows/s each), perform an Interval Join on \texttt{order\_id} matching records where the payment time falls within $\pm$10 minutes of the order time, then aggregate matched records over a 10-minute tumbling window to compute the order count and total payment amount, and output to a console table.

\paragraph{Source Tables.} \texttt{orders\_source} (\texttt{datagen}): \texttt{order\_id}, \texttt{user\_id} in 1--1{,}000{,}000; event time = \texttt{LOCALTIMESTAMP}, 5\,s watermark. \texttt{payments\_source}: \texttt{order\_id} in 1--1{,}000{,}000; \texttt{pay\_amount} in 1--100{,}000; same watermark.

\paragraph{Processing Logic.}
\begin{enumerate}\setlength{\itemsep}{0pt}\setlength{\parsep}{0pt}
\item Define both sources with \texttt{datagen} connectors, \texttt{rows-per-second=1000}.
\item Set \texttt{LOCALTIMESTAMP} as event time with 5\,s watermark on both tables.
\item Interval Join: \texttt{order\_id} match AND payment time within $\pm$10\,min of order.
\item Apply \texttt{TUMBLE} window of 10 minutes over the joined stream.
\item Aggregate: \texttt{COUNT(*) AS order\_count}, \texttt{SUM(pay\_amount) AS total\_pay\_amount}.
\item Convert window start/end to \texttt{VARCHAR}; insert into console output.
\end{enumerate}

\paragraph{Output.} \texttt{console\_output} (print connector): \texttt{window\_start}, \texttt{window\_end}, \texttt{order\_count} (\texttt{BIGINT}), \texttt{total\_pay\_amount} (\texttt{DOUBLE}).

\paragraph{Grading} ($\alpha{=}0.6$): Executability 20, Schema 20, Numerical correctness 45, Configuration 15. No output $\Rightarrow$ all other dims zero. Process 40\%: exploration 35, efficiency 40 (decayed), self-verification 25.

\subsection{Add-Friend Behavior Cluster Aggregation}
\textit{Domain: Security \& Risk Control}
\label{sec:case-security}

\textit{Task ID:} \texttt{hivesql\_014\_en}\hfill\textit{Engine:} HiveSQL

\paragraph{Prompt.} Perform cluster aggregation on add-friend behavior logs to identify malicious clusters. The task processes two filtering branches via \texttt{UNION} and writes aggregated statistics for each cluster to a partitioned table.

\paragraph{Input Table.} Add-friend behavior log with fields \texttt{appname\_}, \texttt{scene\_}, \texttt{headmd5\_}, \texttt{user\_id\_}, \texttt{uinhighquality\_}, and behavioral indicators.

\paragraph{Processing Logic.}
\begin{enumerate}\setlength{\itemsep}{0pt}\setlength{\parsep}{0pt}
\item Time filter: \texttt{day\_} $\in$ [20260608, 20260609], \texttt{hour\_} $\in$ [23, 00].
\item Branch 1: \texttt{appname\_} $\in$ \{hello\_txt, add\_contact\}; sender $\in$ \{CN, HK, MO\}, receiver = CN; \texttt{headmd5\_} non-empty, \texttt{commfrinum\_}=0.
\item Branch 2: \texttt{appname\_}=\texttt{contact\_verify\_ok}; same filters; \texttt{commfrinum\_}=0.
\item Group by \texttt{appname\_}, \texttt{scene\_}, \texttt{headmd5\_}, region and province IDs.
\item Aggregate: \texttt{addfri\_pv}, \texttt{user\_id\_cnt}, quality/evil rates, ID lists via \texttt{collect\_set}.
\item \texttt{HAVING}: \texttt{user\_id\_cnt > 20} AND (\texttt{low\_quality\_rate > 0.99} OR \codeinline{evil\_rate > 0.98}).
\end{enumerate}

\paragraph{Output.} 15-column table via \codeinline{INSERT INTO \dots\ PARTITION}\texttt{(\allowbreak ds=\allowbreak 202606090010)}; grouping keys, PV/UV, quality rates, user ID lists.

\paragraph{Grading} ($\alpha{=}0.5$, Easy): Executability 15, Schema 10, Row alignment 15, Aggregation 40, Write mode 5, Partition 5, UNION branches 10. Process 50\%: exploration 35, efficiency 40 (decayed), self-verification 25.

\balance
\subsection{Code Review Org Ranking}
\textit{Domain: Content, Community \& Dev-Efficiency}
\label{sec:case-content}

\textit{Task ID:} \texttt{pyspark\_016}\hfill\textit{Engine:} PySpark

\paragraph{Prompt.} Generate a PySpark script producing a code-review ranking table organized by four-level organizational hierarchy (company, department, center, group). For each level, rank members by two dimensions---review note count and review time spent (descending, Top 10)---using a cross-join of ranking types and positions to generate a complete ranking space.

\paragraph{Input Tables.} Organization framework, employee-org relation, \texttt{dual} table, CR note user statistics, and CR used-time table (five tables).

\paragraph{Processing Logic.}
\begin{enumerate}\setlength{\itemsep}{0pt}\setlength{\parsep}{0pt}
\item Build four-level org CTEs (company, department, center, group).
\item Associate employees to each level (cp, dp, ct, gp\_user).
\item Compute note counts (by \texttt{author\_id}) and used time (by \texttt{reviewer\_id}).
\item Apply \texttt{rank()} window on two types: notes and used time (desc).
\item Cross-join ranktype $\times$ rank 1--10 with each org level; left-join results.
\item Union all four levels; write to target table.
\end{enumerate}

\paragraph{Output.} 17-column table via \codeinline{INSERT OVERWRITE TABLE}; \texttt{org\_id}, \codeinline{org\_type}, \texttt{ranktype}, \texttt{rank}, \texttt{user\_id}, value columns. ORC format.

\paragraph{Grading} ($\alpha{=}0.7$): Executability 10, Hierarchy completeness 20, Ranking correctness 25, Completeness 25, Schema \& storage 20. Process 30\%: exploration 35, efficiency 40 (decayed), self-verification 25.

\subsection{Ad ASA Attribution Migration}
\textit{Domain: Advertising \& Marketing}
\label{sec:case-advertising}

\textit{Task ID:} \texttt{prestosql\_002\_en}\hfill\textit{Engine:} PrestoSQL/Trino

\paragraph{Prompt.} From ad attribution raw logs, filter valid records of two attribution types (\texttt{ad\_attribution} and \texttt{phonebrandh}), aggregate the log count and deduplicated user count by version and media dimensions, and write results using Presto/Trino SQL syntax (no Hive/Spark dialects).

\paragraph{Input Table.} Ad attribution raw log with 27 fields including \texttt{ad\_version}, \texttt{attribution}, \texttt{callback}, \texttt{device\_id}, and partition \texttt{ds}.

\paragraph{Processing Logic.}
\begin{enumerate}\setlength{\itemsep}{0pt}\setlength{\parsep}{0pt}
\item Filter: \texttt{ds = '2026060919'}.
\item Valid: (\texttt{ad\_attribution} + \texttt{attribution=\allowbreak 'true'}) OR (\texttt{phonebrandh} + \texttt{callback} non-empty).
\item Map \texttt{media\_id}: \texttt{ad\_attribution}$\to$39, \texttt{phonebrandh}$\to$28.
\item Cast \texttt{ds} to \texttt{BIGINT}; aggregate \texttt{log\_num=COUNT(*)}.
\item Deduplicate: \texttt{log\_num\_qimei=COUNT(DISTINCT device\_id)}; group by \texttt{ds}, \texttt{ad\_version}.
\end{enumerate}

\paragraph{Output.} 5-column table: \texttt{ds} (\texttt{BIGINT}), \texttt{ad\_version} (\texttt{STRING}), \texttt{media\_id} (\texttt{INT}), \texttt{log\_num} (\texttt{BIGINT}), \texttt{log\_num\_qimei} (\texttt{BIGINT}). Execute via \codeinline{presto-cli}.

\paragraph{Grading} ($\alpha{=}0.6$, Medium): Executability 15, Schema 10, Row alignment 15, Attribution filter 10, Media ID 10, Aggregation 15, Completeness 5, Write mode 5, Partition 5, Source filter 10. Process 40\%: exploration 35, efficiency 40 (decayed), self-verification 25.

\endgroup

\end{document}